\documentclass[lettersize,journal]{IEEEtran}
\usepackage{amsmath,amsfonts}
\usepackage{algorithmic}
\usepackage{algorithm}
\usepackage{array}
\usepackage[small]{caption}
\usepackage{subcaption}
\usepackage{textcomp}
\usepackage{stfloats}
\usepackage{url}
\usepackage{verbatim}
\usepackage{graphicx}
\usepackage{cite}

\usepackage{cleveref}
\usepackage{xspace}
\usepackage{multirow}
\usepackage{multicol}
\usepackage{makecell}
\usepackage{enumitem}
\usepackage{xspace}
\usepackage{amssymb}
\usepackage{amsthm}
\usepackage{booktabs}
\makeatletter
\DeclareRobustCommand\onedot{\futurelet\@let@token\@onedot}
\def\@onedot{\ifx\@let@token.\else.\null\fi\xspace}

\def\eg{\emph{e.g}\onedot} 
\def\ie{\emph{i.e}\onedot} 
 
\def\etc{\emph{etc}\onedot} \def\vs{\emph{vs}\onedot}

\makeatother

\newtheorem{theorem}{Theorem}
\newtheorem{definition}{Definition}

\begin{document}

\title{Disentangled Noisy Correspondence Learning}

\author{Zhuohang Dang, Minnan Luo$^*$, Jihong Wang, Chengyou Jia, Haochen Han, Herun Wan\\ Guang Dai, Xiaojun Chang, and Jingdong Wang
\thanks{Corresponding author: Minnan Luo.}
\thanks{Zhuohang Dang, Minnan Luo, Jihong Wang, Chengyou Jia, Haochen Han and Herun Wan are with the School of Computer Science and Technology, the Ministry of Education Key Laboratory of Intelligent Networks and Network Security, and the Shaanxi Province Key Laboratory of Big Data Knowledge Engineering, Xi’an Jiaotong University, Xi’an, Shaanxi 710049, China.
{e-mail: \{dangzhuohang, wang1946456505, cp3jia, hhc1997, wanherun\}@stu.xjtu.edu.cn, minnluo@xjtu.edu.cn}.}
\thanks{Guang Dai is with SGIT AI Lab, and also with State Grid Shaanxi Electric Power Company Limited, State Grid Corporation of China, Shaanxi, China, {e-mail: guang.gdai@gmail.com}.}
\thanks{Xiaojun Chang is with the School of Information Science and Technology, University of Science and Technology of China. Xiaojun Chang is also a Visiting Professor with Department of Computer Vision, Mohamed bin Zayed University of Artificial Intelligence (MBZUAI). {e-mail: cxj273@gmail.com.}}
\thanks{Jingdong Wang is with the Baidu Inc, China, {e-mail: wangjingdong@outlook.com.}}
}

\markboth{Journal of \LaTeX\ Class Files,~Vol.~14, No.~8, August~2021}
{Shell \MakeLowercase{\textit{et al.}}: A Sample Article Using IEEEtran.cls for IEEE Journals}

\maketitle

\begin{abstract}

  Cross-modal retrieval is crucial in understanding latent correspondences across modalities.
  However, existing methods implicitly assume well-matched training data, which is impractical as real-world data inevitably involves imperfect alignments, \ie, noisy correspondences.
  Although some works explore \textbf{similarity-based strategies} to address such noise, they suffer from sub-optimal similarity predictions influenced by modality-exclusive information (MEI), \eg, background noise in images and abstract definitions in texts.
  This issue arises as MEI is not shared across modalities,  thus aligning it in training can markedly mislead similarity predictions.
  Moreover, although intuitive, directly applying previous cross-modal disentanglement methods suffers from limited noise tolerance and disentanglement efficacy.
  Inspired by the robustness of information bottlenecks against noise, we introduce \textbf{DisNCL}, a novel information-theoretic framework for feature \textbf{Dis}entanglement in \textbf{N}oisy \textbf{C}orrespondence \textbf{L}earning, to adaptively balance the extraction of MII and MEI with certifiable optimal cross-modal disentanglement efficacy.
  DisNCL then enhances similarity predictions in modality-invariant subspace, thereby greatly boosting similarity-based alleviation strategy for noisy correspondences.
  Furthermore, DisNCL introduces soft matching targets to model noisy many-to-many relationships inherent in multi-modal input for noise-robust and accurate cross-modal alignment.
  Extensive experiments confirm DisNCL's efficacy by 2\% average recall improvement. Mutual information estimation and visualization results show that DisNCL learns meaningful MII/MEI subspaces, 
  validating our theoretical analyses.
  \end{abstract}


\section{Introduction}

Cross-modal retrieval \cite{diao2021similarity, huang2018bi, zhang2022latent} aims to retrieve the most relevant samples from different modalities, crucial for various domains such as criminal investigation  \cite{gao2022conditional}. 
Existing methods typically first project multi-modal inputs to a unified feature space. 
They then devise feature interaction strategies at either global \cite{chen2021learning,frome2013devise,zhang2020deep} or local \cite{lee2018stacked,diao2021similarity,diao2023plug} levels for similarity prediction, enhancing similarities for matching pairs while suppressing similarities of mismatched ones.
Although effective, a core assumption of these methods is well-matched train data, which is inconsistent with real-world data.
For example, Conceptual Captions \cite{sharma2018conceptual} collects 3.3 million co-occurring sample pairs from the Internet, where around 20\% samples are mismatched \cite{alayrac2022flamingo,huang2021learning}, \ie, noisy correspondences.
\Cref{table:flicker,table: CLIP comparison} demonstrate that the traditional training paradigm, whether training from scratch or fine-tuning pretrained models like CLIP \cite{radford2021learning}, significantly underperforms on such noisy data, necessitating robust learning strategies for noisy correspondences.

\begin{figure}[!t]
  \centering
  \subfloat[Entangled noisy correspondence learning methods.]{\includegraphics[width=1\linewidth]{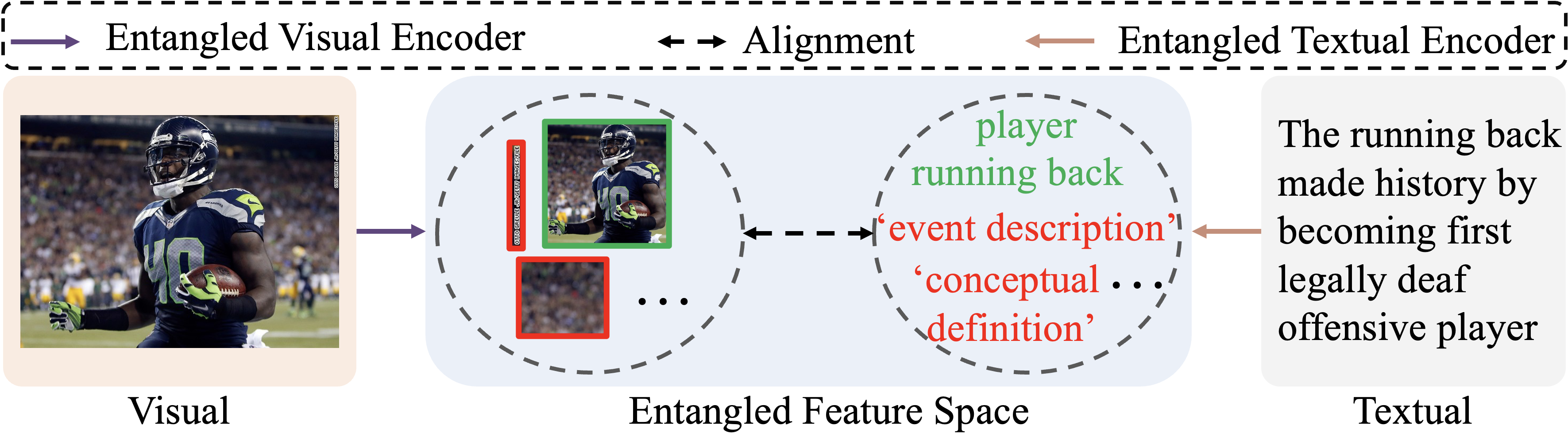}\label{fig: intro_entangled}}
  \\
  \subfloat[The proposed DisNCL.]{\includegraphics[width=1\linewidth]{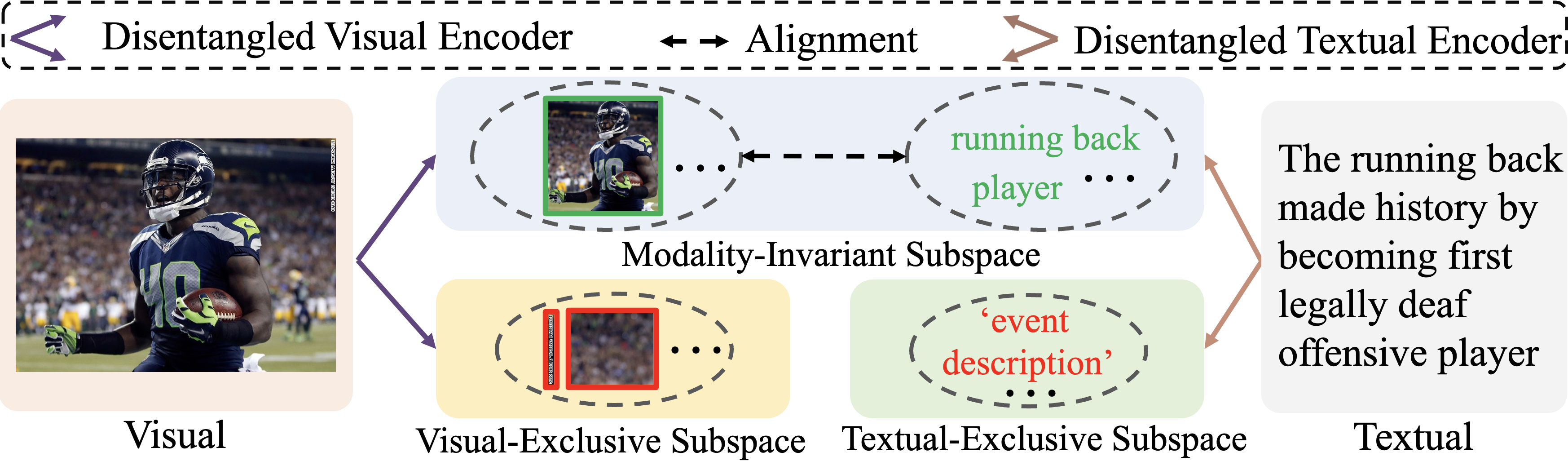}\label{fig: intro_disentangled}}
\caption{Illustrative comparisons of entangled methods and our DisNCL, where different colors indicate corresponding feature space. The red/green elements refer to MII and MEI, respectively.
The event description and conceptual definition in textual MEI refers to `history' and `legally deaf' in text.
}
\vspace{-4mm}
\end{figure}

In this context, recent works introduce noise during training to bridge the gap between well-matched datasets and noisy real-world data for enhanced robustness.
Typically, these methods \cite{han2023noisy,yang2023bicro,huang2021learning} employ \textbf{similarity-based strategies for identification and suppression of these mismatched pairs.}
Specifically, they first employ the memorization effect of deep neural networks \cite{arpit2017closer,arazo2019unsupervised}, which enables clean samples to exhibit higher similarities than noisy ones after the initial few epochs \cite{yao2020searching}.
Then they use samples' similarity distribution, showing a clear bimodal distribution, to coarsely identify and filter out noise data. 
Subsequently, they derive a per-sample adaptive margin from its similarity prediction to replace the constant scalar margin in triplet ranking loss \cite{hermans2017defense,weinberger2009distance}.
This adaptive margin facilitates potential noisy correspondences' compliance with the triplet ranking loss, effectively excluding the risky supervision information during training.

However, previous methods suffer from two significant challenges.
\textbf{1) Feature Entanglement}: \Cref{fig: intro_entangled} shows these methods predict similarities in a unified feature space with modality-invariant information (MII) and modality-exclusive information (MEI) entangled.
Specifically, MII captures shared components across modalities, \eg, `the running back', serving as foundation for retrieval.
Conversely, MEI is not shared across modalities, \eg, background noise, watermark in image and abstract descriptions such as `legally deaf' and `history' in text.
Consequently, aligning MEI across modalities can erroneously skew similarity predictions towards sub-optimality, misleading similarity-based identification and suppression of noisy correspondences.
Although some works introduce various strategies, \eg, feature consistency \cite{yang2022disentangled} and cross prediction \cite{xia2023achieving}, to disentangle MII from MEI, they fall short in noise tolerance and disentanglement efficacy. 
Specifically, for optimal MII extraction, these methods intuitively minimize the distance between MII or exchange MII to reconstruct input from another modality, which are \emph{sensitive to incorrect alignment information from noisy correspondences and may converge trivially to a subset of optimal MII while leaking the remainder to MEI.}
\textbf{2) Exclusive Correspondence:}  the hard negative strategy \cite{faghri2017vse++} in previous methods \cite{han2023noisy,yang2023bicro} can only model the one-to-one correspondence between given query and its positive or hard negative samples (\Cref{fig: soft_similarity}).
This strategy is sub-optimal as it fails to consider the noisy many-to-many correspondences inherent in multi-modal data \cite{andonian2022robust, morgado2021robust}, \eg, shared entities among unpaired images and texts.

In light of above, we introduce \textbf{DisNCL}, a pioneering information-theoretic framework for feature \textbf{Dis}entanglement in \textbf{N}oisy \textbf{C}orrespondence \textbf{L}earning, to enhance models' robustness against mismatched pairs in training. 
Specifically, we introduce a novel objective based on information bottleneck principle \cite{alemi2016deep} to extract two complementary components, \ie, MII and MEI.
Subsequently, as illustrated in \Cref{fig: intro_disentangled}, we conduct similarity predictions within the disentangled modality-invariant subspace, together with the cross-modal retrieval training. 
This methodology aims to direct the model's focus solely on MII and exclude the detrimental impacts of MEI, boosting identification and negative impact suppression of mismatched samples with more accurate similarity predictions.
Furthermore, we estimate softened targets with bootstrapping strategy in the modality-invariant subspace for sample matching.
These softened targets, contrasting with one-to-one correspondence in hard negative strategy, enable more complex many-to-many cross-modal relationship modeling to further enhance model's efficacy. 
Moreover, it is noteworthy that we theoretically prove the optimal disentanglement of MII and MEI, along with the noise robustness of our soft cross-modal alignment, confirming our DisNCL's superior efficacy.

Our contribution is summarized as follows:
\begin{itemize}
    \item We introduce DisNCL, the first work to introduce certifiable optimal cross-modal disentanglement efficacy for noisy correspondence learning, for enhanced robustness against noisy multi-modal training data.
    \item We boost identification and suppression of noisy correspondences by accurate similarities predicted in modality-invariant subspace. 
    Moreover, we propose a noise-robust alignment refinement strategy to model the intricate relationships in multi-modal data via soft target estimation.
    \item Extensive experiments on various benchmarks confirm DisNCL’s efficacy by improving 2\% average recall, as well as validate our theoretical analyses.
\end{itemize}

The remainder of the article is organized as follows. In Section \uppercase\expandafter{\romannumeral2}, we briefly summarize the related works, \eg, noisy correspondence learning and cross-modal disentanglement. In
    Section \uppercase\expandafter{\romannumeral3}, we describe the proposed DisNCL in detail. 
    In Section \uppercase\expandafter{\romannumeral4}, we derive variational estimations of our DisNCL for stable optimization.
    In Section \uppercase\expandafter{\romannumeral5}, we provide in-depth theoretical analyses of DisNCL's disentanglement efficacy and noise robustness.
    In Section \uppercase\expandafter{\romannumeral6}, we firstly present datasets and implementation details. We then report experimental results comparing with SOTA methods. We also conduct the ablation study and provide more detailed analyses. 
    Section \uppercase\expandafter{\romannumeral7} concludes this paper and describes the future work.

\section{Related Work}
\subsection{Noisy Correspondence Learning}

Contemporary cross-modal retrieval methods primarily diverge into global-level methods \cite{radford2021learning,chun2021probabilistic,chen2021learning} that treat images and sentences as holistic entities and local-level \cite{lee2018stacked,pan2023fine,diao2021similarity,jing2021learning} ones focusing on essential fragments \eg, salient objects in images and keywords in texts. 
However, these methods all assume well-matched training data, inconsistent with real-world data and thus degrading performance.

Specifically, due to lack of manual annotation, web-crawled datasets, \eg, Conceptual Captions \cite{sharma2018conceptual} and M3W \cite{alayrac2022flamingo}, inevitably contain noisy correspondences (NCs) \cite{huang2021learning}.
This issue has been extensively explored in various domains, including  person re-id \cite{yang2022learning}, multi-view learning \cite{yang2022robust,yang2021partially}, and graph matching \cite{lin2023graph}, \etc
In image-text matching, RCL \cite{hu2023cross} and DECL \cite{qin2022deep} propose novel losses, \eg, complementary contrastive learning and dynamic hardness, to suppress the amplification of wrong supervision inherent in NCs. 
However, they lack explicit strategies for easily separable noisy samples \cite{qin2023cross},  leading to limited efficacy.
To this issue, NCR \cite{huang2021learning} leverages the memorization effect of neural networks to identify NCs in a co-teaching manner.  MSCN \cite{han2023noisy} employs meta-learning for enhanced similarity predictions. BiCro \cite{yang2023bicro} explores bidirectional similarity consistency to derive more accurate adaptive margins.
Moreover, CRCL \cite{qin2023cross} employs exponential normalization to aggregate historical similarity predictions for more stable suppression of NCs.
Despite their advancements, these methods suffer from sub-optimal similarity predictions in entangled feature space. In contrast, we use cross-modal disentanglement for accurate similarity prediction in a modality-invariant space, effectively eliminating the adverse effects of modality-exclusive noise.

\begin{figure*}[!t]
  \centering
  \includegraphics[width=1\textwidth]{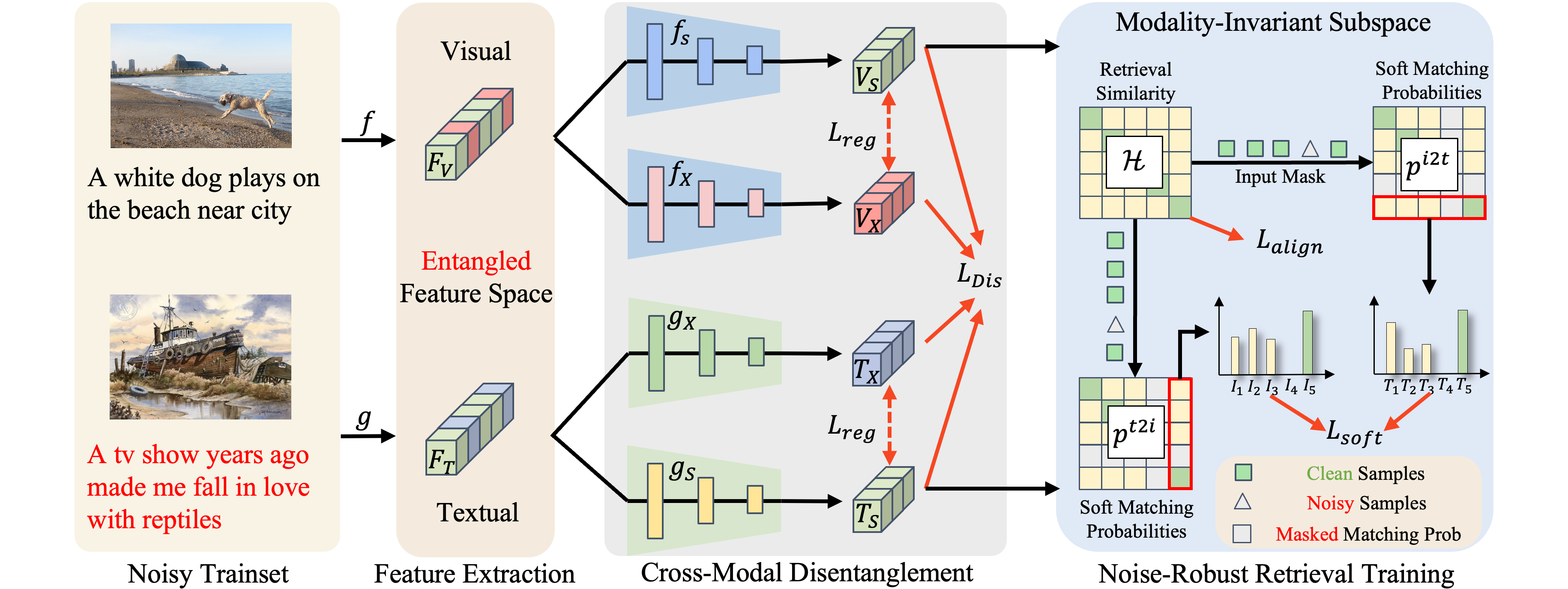}
\caption{The overview of our DisNCL, where black and red arrows indicate the model forward and optimization constraints; the green blocks indicate MII, while the pink and blue ones denote MEI of $V$ and $T$, respectively.}
\label{fig: overview}
\vspace{-6mm}
\end{figure*}

\subsection{Cross-Modal Disentanglement}

Prevailing methods typically employ objectives with intuitive or heuristic constraints to extract complementary modality-invariant and modality-exclusive information from data.
Representatively,  DMIM \cite{guo2019learning}, CMG \cite{xia2023achieving} and MDVAE \cite{tian2022multimodal} integrate cross prediction strategy, focusing on extracting identical modality-unified (invariant) information.
Additionally, FDMER \cite{yang2022disentangled} introduces feature consistency constraints to ensure commonality across the modality-invariant representations. 
However, these methods cannot ensure the optimal modality-invariant representations, \eg, they may learn only a subset of the optimal modality-invariant information while leaking the remainder to modality-exclusive information.
Moreover, these strategies to directly force the feature invariance are sensitive to wrong alignment of noisy correspondences in data.
On the other hand, information regularization \cite{mao2021deep,tian2021farewell} methods employ extra regularizers to preserve modality-invariant information while discarding modality-exclusive noise, showcasing improved noise tolerance.
However, it still fails to ensure optimal disentanglement efficacy due to inevitable performance-compression trade-off of trivial regularizers \cite{pan2021disentangled, dang2023disentangled,dang2024disentangled}, \ie, the performance decreases as the level of compression intensifies.
In contrast, with a rigorous definition of modality-exclusive and modality-invariant information, our DisNCL demonstrates certifiable optimal cross-modal disentanglement efficacy and appealing noise tolerance highlighted in \Cref{th: theoretical analyses,th: noise robust}.

\section{Methodology}

\subsection{Problem Definition}

We follow \cite{huang2021learning,han2023noisy} to use the image-text retrieval task as a proxy to investigate the noisy correspondence problem in cross-modal retrieval.
Specifically, let $\mathcal{D} = {(V_i, T_i)}_{i=1}^N$ denote a well-matched image-text dataset with $N$ training samples.
In noisy correspondence learning, an unknown proportion of supposed positive pairs in $\mathcal{D}$ are mismatched, \ie, the image and text do not correspond, despite they belong to $\mathcal{D}$ as in \Cref{fig: overview}. This misalignment introduces erroneous supervisory information, which can mislead the model during training and remarkably degrade performance. 

\subsection{Model Overview}
This section outlines our DisNCL's detail, with an overview shown in \Cref{fig: overview}.
We first use feature encoders to extract original representations from input pairs.
Then we advance the information bottleneck theory by introducing a novel information-theoretic objective $L_{Dis}$ and $L_{reg}$ for extracting modality-invariant information (MII) and modality-exclusive information (MEI), respectively.
Next, within modality-invariant subspace, we conduct similarity computation and retrieval training, where softened targets, instead of original hard matching labels, are estimated for each sample pair to facilitate more accurate cross-modal alignment.
We will detail each component and its optimization objective in what follows.

\subsection{Cross-Modal Information Disentanglement}

Given a set of visual-textual pairs $\{{V_i,T_i}\}_{i=1}^{N}$, we first use modality-specific encoders $f$ and $g$ to encode input into a unified feature space, \ie, $F_V^i=f(V_i)$ and $F_T^i=g(T_i)$, respectively.
Next, we aim to disentangle MII from MEI within multi-modal input pairs, thereby excluding the adverse effects of MEI on similarity predictions. 
Specifically, let $f_S(\cdot), f_X(\cdot)$ be the disentangled encoders for visual, and $g_S(\cdot), g_X(\cdot)$ for textual. 
We seek a disentangled representation space $F_V=(V_S,V_X)$ and $F_T=(T_S,T_X)$, where the MII is encoded as $V_S=f_S(F_V)$ and $T_S=g_S(F_T)$, while the MEI is captured by $V_X=f_X(F_V)$ and $T_X=g_X(F_T)$, respectively. 
Then, we employ the information theory to define the desired MII and MEI in \Cref{th: factorization} by quantifying the input information within corresponding representations \cite{guo2024learning}. 
Crucially, this formulation helps us identify the optimal information-theoretic constraints for MII and MEI, paving the way for our theoretical analyses of cross-modal disentanglement efficacy in \Cref{th: theoretical analyses}.
\begin{theorem}
  Given a multi-modal input pair $(V,T)$ with corresponding representations $F_V$ and $F_T$, the mutual information $I(T; F_T)$ and $I(V; F_V)$  can be decomposed into two complementary terms, i.e.,
  \begin{equation*}
  \begin{aligned}
  &I(T; F_T) &= &\quad\quad I(V; F_T) &+ &\quad\  I(T; F_T|V) &,\\
    &I(V; F_V) &= &\underbrace{I(T; F_V)}_{modality-invariant} &+ &\underbrace{I(V; F_V|T)}_{modality-exclusive} &.
  \end{aligned}
  \end{equation*}
  \label{th: factorization}
  \begin{proof}
    Please refer to Appendix A-A for detailed proof.
  \end{proof}
  \end{theorem}
Representatively, $I(T; F_V)$ is the MII term since it quantifies the consistent visual information in $F_V$ across $T$.
In contrast, $I(V; F_V|T)$ is the MEI term since it captures the unique modality variance in $V$ but absent from $T$.
Building upon \Cref{th: factorization}, we introduce a novel information-theoretic objective on filtered clean data $\mathcal{D}_{\text{clean}}$ (\Cref{eq: filter}) to discern and capture MII and MEI from multi-modal inputs, \ie,
\begin{definition}
  For a multi-modal input pair $(V,T)$ with feature $(F_V,F_T)$, the desired disentangled representations $F_V=(V_S,V_X), F_T=(T_S,T_X)$ can be achieved by solving optimization objective:
  \begin{equation}
    \min L_{Dis}=\gamma L_S + (1-\gamma)L_X,
  \end{equation}
  where $\gamma$ is a hyperparameter that controls the tradeoff between two objectives; $L_S$ and $L_X$ denote the modality-invariant objective and modality-exclusive objective, \ie:
  \begin{itemize}[leftmargin=3mm]
    \item $L_S$: DisNCL learns MII by maximizing modality-invariant term while minimizing modality-exclusive term, \ie,
    {\small
    \begin{equation*}
        L_S=-(I(T; V_S)+I(V;T_S))+\beta_1(I(V; V_S|T)+I(T; T_S|V)).
    \end{equation*}
    }
    \item $L_X$: DisNCL learns MEI by maximizing modality-exclusive term while minimizing modality-invariant term, \ie,
      {\small
        \begin{equation*}
        L_X=-(I(V; V_X|T)+I(T; T_X|V))+\beta_2(I(T; V_X)+I(V; T_X)).
    \end{equation*}}
  \end{itemize}
  \label{def: objective}
  \end{definition}
Here $\beta_1$ and $\beta_2$ are two pre-defined hyperparameters. As in Appendix A-B, $L_S$ and $L_X$ are equivalent to information bottlenecks (IBs) \cite{alemi2016deep}, preserving desirable information while discarding irrelevant information in representations. In this sense, $L_{Dis}$ is totally information theoretic based, complying with the desired information extraction and preservation in cross-modal disentanglement. 
Moreover, its information bottleneck based objective formulation can maintain high noise tolerance for enhanced model robustness \cite{wan2021multi,wei2022contrastive}.

\noindent\textbf{Theoretical Insight.}
By viewing images and texts as two views of an entity, our $L_{Dis}$ is reduced to \cite{federici2020learning} when $\gamma=1$, which only extracts modality-invariant (view-shared) information during training. 
However, there are three fundamental differences between ours and \cite{federici2020learning}: 1) We integrate the extraction of MII and MEI into a unified framework, showcasing better generalizability in tasks that require MEI to explore modality complementarity, \eg, sentiment analysis and autonomous driving. 2) \cite{federici2020learning} relies on the strong assumption that each view/modal provides the same task-relevant information. 
Instead, we explore the noisy correspondence scenario where information between views may be mismatched. 
In this sense, we further highlight our model's noise robustness in \Cref{th: noise robust} that our model trained on such noisy data is equivalent to that trained on clean data, demonstrating our DisNCL's superior efficacy.
3) Crucially, we prove in \Cref{sec: theoretical} to highlight DisNCL’s optimal cross-modal disentanglement efficacy, validated by ablation study in \Cref{sec: disentanglement} and therefore outperform \cite{federici2020learning} by a notable margin in \Cref{sec: disentanglement comparsion}.

\subsection{Sample Filtration and Robust Hinge Loss}
As mentioned above, we train encoders with $L_{Dis}$ to obtain the disentangled cross-modal representations.
Next, we predict similarities within modality-invariant subspace to mitigate MEI's adverse effects.
Specifically, given a batch of multi-modal input pairs ${(V_i,T_j)}_{i,j=1}^B$ with batch size $B$, while $V_S^i=f_S(f(V_i))$ and $T_S^j=g_S(g(T_j))$ are modality-invariant representations.
We employ a function $\mathcal{H}$ to compute similarity for $(V_i,T_j)$ as $\mathcal{H}(V_S^i,T_S^j)$, where $\mathcal{H}$ can be a cosine function \cite{faghri2017vse++,chen2021learning} or learnable module \cite{huang2021learning,han2023noisy}.
For brevity, $\mathcal{H}(V_S^i,T_S^j)$ is denoted as $\mathcal{H}_{ij}$ or $\mathcal{H}(V_i,T_j)$ in the following.

\begin{figure}[!t]
  \includegraphics[width=1\linewidth]{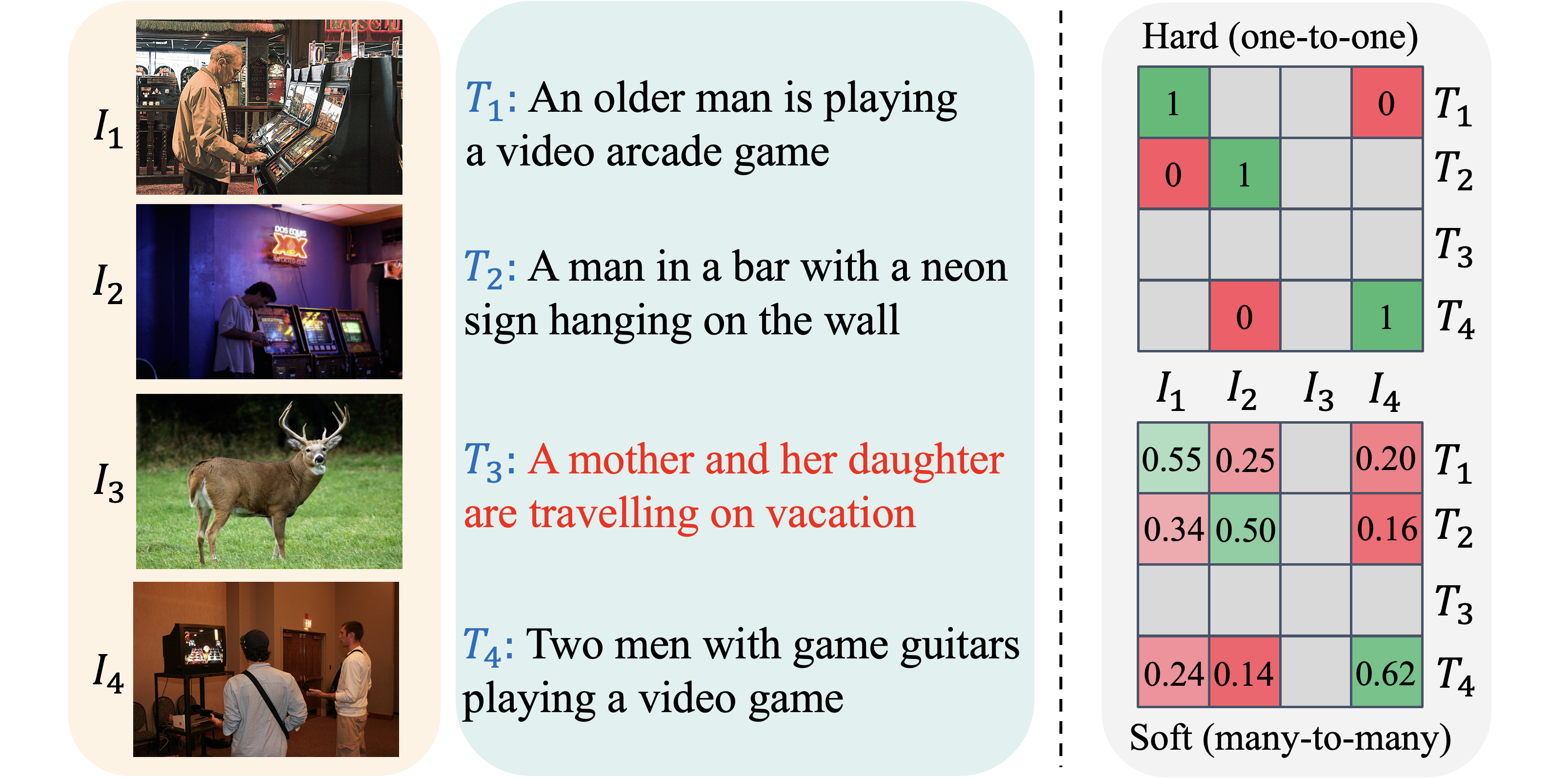}
  \caption{Illustration of hard negative strategy's one-to-one (above) and our soft many-to-many correspondence (below), where gray blocks denote the masked pairs, and green/red blocks indicate positive/negative samples.}
  \label{fig: soft_similarity}
  \vspace{-4mm}
\end{figure}

Subsequently, we follow \cite{qin2022deep} to leverage the maximum similarity constraints within the batch to identify potential noisy  correspondences, \ie,
\begin{equation}
\mathcal{D}_{\text{clean}} = \{i\mid i=\arg\textstyle{\max_j}\mathcal{H}_{ij}\ \text{and} \ \ i=\arg\textstyle{\max_j}\mathcal{H}_{ji}\},
\label{eq: filter}
\end{equation}
which filters out samples with off-diagonal argmax outputs, as potential matches in retrieval tasks are expected to appear along the diagonal.
Then, a hinge-based ranking loss with an adaptive soft margin is employed for robust training, \ie,
\begin{equation}
  \begin{aligned}
    L_{align}(I_{i}, T_{i})  =&\sum_{i\in\mathcal{D}_{\text{clean}}}\left\{[\hat{\alpha}_{i}-\mathcal{H}(I_{i}, T_{i})+\mathcal{H}(I_{i}, \hat{T}_{h})]_{+}\right. \\
    +&\left.[\hat{\alpha}_{i}-\mathcal{H}(I_{i}, T_{i})+\mathcal{H}(\hat{I}_{h}, T_{i})]_{+}\right\},
    \end{aligned}
\label{eq: alignment}
\end{equation}
where $[x]_+=\max(x,0)$.
Following \cite{han2023noisy,yang2023bicro}, $\hat{\alpha}_{i}=\frac{m^{\mathcal{H}_{ii}}-1}{m-1}\alpha$ is the soft margin for the $i$-th sample pair with a constant margin $\alpha$ and hyperparameter $m$. 
$\hat{I}_{h}$ and $\hat{T}_{h}$ are the hard negative samples for text and image, respectively, \eg, $\hat{T}_{h}$ is the negative text most similar to $I_i$ within the batch.
Notably, \Cref{eq: filter,eq: alignment} are based on more accurate similarity predictions within modality-invariant subspace, paving the way for more effective identification and negative impact suppression of mismatched pairs, thereby greatly enhancing robustness in noisy correspondence learning.

\subsection{Softened Cross-Modal Alignment}
However, as shown in \Cref{fig: soft_similarity}, the hard negative strategy in \Cref{eq: alignment} is constrained to modeling the one-to-one correspondence between given query and its positive (green diagonal) or hard negative (red) samples.
This strategy is sub-optimal, especially considering that the image-text relationship is not a strict one-to-one but rather a noisy many-to-many correspondence \cite{gao2023softclip,srinivasa2023cwcl}.
Specifically, there can be local similarities among negative image-text pairs within a batch, \eg, video game in $I_1,I_2,I_3$ and $T_1,T_4$, indicating a nuanced, multi-faceted relationship.
As a result, we employ soft targets for more accurately modeling such complex relationships.

Considering the inherent consistency across different modalities in matched data, the similarity in one modality can naturally serve as softened targets for the other to aggregate information over all other instances from the opposite modality \cite{reed2014training,morgado2021robust,andonian2022robust}, therefore enabling a more refined cross-modal alignment.
Notably, \textbf{such consistency does not hold for noisy correspondence as it is mismatched}. Therefore, we first exclude them from the soft target estimation.
Specifically, let $\mathcal{H}^{i2t}=\mathcal{H}$ and $\mathcal{H}^{t2i}=\mathcal{H}^T$, we have:
\begin{equation}
  \forall i\notin\mathcal{D}_{\text{clean}},\quad j\in[1,B],\quad \mathcal{H}^{i2t}_{ij}=\mathcal{H}^{t2i}_{ij}=0.
\end{equation}
Then we follow \cite{yang2021partially} to derive the bidirectional matching probabilities for a pair $(V_i,T_j)$ as follows:
\begin{equation}
  p^{i2t}_{ij}=\frac{e^{\mathcal{H}^{i2t}_{ij}/\tau}}{\sum_{l=1}^Be^{\mathcal{H}^{i2t}_{il}/\tau}},\quad\quad
  p^{t2i}_{ij}=\frac{e^{\mathcal{H}^{t2i}_{ij}/\tau}}{\sum_{l=1}^Be^{\mathcal{H}^{t2i}_{il}/\tau}},
  \label{eq: soft logits}\end{equation}
where $\tau$ is a temperature parameter.
Crucially, the retrieval logits in \Cref{eq: soft logits} are derived in modality-invariant subspace, thereby avoiding the impact of modality-exclusive noise and accurately reflecting the given sample's similarity with others across modalities.
Subsequently, we use bootstrapping strategy \cite{reed2014training} to formulate the soft alignment loss, \ie,
  \begin{align}
    L_{soft}=\sum_{i\in\mathcal{D}_{\text{clean}}}& H(p^{i2t}_i, y_i^{i2t}) + H(p^{t2i}_i, y_i^{t2i}), 
    \label{eq: soft alignment}
  \end{align}
where $H(\cdot,\cdot)$ is the entropy function; $y_i^{i2t} = \beta_3\mathbf{e}_i +(1-\beta_3)p^{t2i}_i$ and $y_i^{t2i} = \beta_3\mathbf{e}_i +(1-\beta_3)p^{i2t}_i$ is the bootstrapping soft label of the $i$-th sample, where $\mathbf{e}_i$ is the one-hot label with the $i$-th element being 1 and $\beta_3$ is a hyperparameter for controlling the tradeoff between noise robustness and soft alignment.

\textbf{Theoretical Insight}: Our $L_{soft}$ achieves fine-grained cross-modal alignment by viewing sample matching as a classification task within batch. Specifically, $\mathcal{H}^{i2t}$ and $\mathcal{H}^{t2i}$ can be viewed as image-to-text and text-to-image classification logits, with $p^{i2t}$ and $p^{t2i}$ as corresponding probabilities. Therefore, the model can be formulated as an ensemble of two classifiers, $f_{i2t}(I_i,T) = \mathcal{H}^{i2t}_i$ and $f_{t2i}(I,T_i) = \mathcal{H}^{t2i}_i$.
In this sense, $\mathbf{e}_i$ is a noisy label for the $i$-th sample within batch, as the noisy correspondences in data will indicate wrong sample classification target, thereby misleading model optimization towards suboptimality.
To this issue, we demonstrate the noise robustness of $f_{i2t}$ and $f_{t2i}$ with $L_{soft}$ in \Cref{th: noise robust}, thereby effectively achieving fine-grained soft cross-modal alignment while ensuring model's robustness to noisy correspondence.

\subsection{Training Objective}
To prevent sub-optimal solutions, we follow \cite{pan2021disentangled} to further employ a regularizer to encourage the disentanglement of modality-invariant and modality-exclusive representations, \ie,
\begin{equation}
  L_{reg}=I(V_S;V_X)+I(T_S;T_X).
\end{equation}
Subsequently, the final training objective is formulated as:
\begin{equation}
  \min L_{DisNCL}=L_{Dis}+L_{align}+L_{soft}+L_{reg}.
\end{equation}
Note that $L_{Dis}$ and $L_{reg}$ are intractable to directly optimize since their mutual information terms consist of integral on high-dimensional data. To this issue, we design tractable optimization strategy for stable optimization of these terms in what follows.

\section{DisNCL Optimization}
\label{sec: estimation}
In this section, we derive variational estimations of $L_{Dis}$ ($L_S$ and $L_X$) and $L_{reg}$ for stable optimization of their mutual information objectives.
\subsection{Estimation of $L_S$}
\subsubsection{Maximization of $I(T; V_S)$ and $I(V; T_S)$}
We assume that all information in $T$ is captured by its representation $F_T$, \ie, $I(T; V_S)=I(F_T,V_S)$, and directly use the Jensen-Shannon estimator \cite{nowozin2016f} to maximize $I(F_T; V_S)$ as an alternative, formulated as:
  \begin{align}
    \max _{V_S, F_T, f_{1}} I_{J S D}\left(V_S ; F_T\right)&=  \mathbb{E}_{p\left(V_S\right)p\left(F_T\right)}\left[\log \left(1-f_{1}\left(V_S, F_T\right)\right)\right] \notag\\
    & +\mathbb{E}_{p\left(V_S, F_T\right)}\left[\log f_{1}\left(V_S, F_T\right)\right] , \label{eq: L_S}
    \end{align}
where $f_1$ is a learnable discriminator to discriminates whether $F_T$ and $V_S$ are correlated. The $I(V; T_S)$ can be optimized similarly.
\subsubsection{Minimization of $I(V; V_S|T)$ and $I(T; T_S|V)$}
We derive a tra-ctable upper bound for minimizing $I(V; V_S|T)$ as:
\begin{equation}
  I(V; V_S|T)\leq KL(p(V_S|V)||p(T_S|T)),
\end{equation}
where $p(V_S|V)$ and $p(T_S|T)$ is the disentangled modality-invariant information extraction (see Appendix A-D for details).
We assume these extractions follow a Gaussian distribution, \eg, $p(V_S|V)\thicksim\mathcal{N}(\mu(V);\textbf{I})$ with fixed variance and $\mu=f_S\circ f$ is the composition of $f_S$ and $f$. The $I(T; T_S|V)$ can be optimized in a similar way.

\subsection{Estimation of $L_X$ and $L_{reg}$}
For simplicity of optimization, we first reformulate modality-exclusive objective $L_X$ (see Appendix A-B for details), \ie,
\begin{align}
  L_X=&-(I(V; V_X)+I(T; T_X)) \notag\\
  &+(1+\beta_2)(I(T; V_X)+I(V; T_X)),
\end{align}
where $I(V; V_X)$ and $I(T; T_X)$ can be maximized similarly as $I(T; V_S)$.
Moreover, as $I(T;V_X)=I(F_T,V_X)$,  we follow \cite{pan2021disentangled} to adopt an adversarial strategy to minimize $I(F_T,V_X)$ as an alternative, \ie, 
  \begin{align}
    \min_{V_X,F_T}\max _{f_{2}} I\left(V_X ; F_T\right)&=  \mathbb{E}_{p\left(V_X\right)p\left(F_T\right)}\left[\log (1 - f_{2}\left(V_X, F_T\right))\right] \notag\\
    & +\mathbb{E}_{p\left(V_X, F_T\right)}\left[\log f_{2}\left(V_X, F_T\right)\right], \label{eq: L_X}
    \end{align}
where $f_2$ is a  discriminator to discriminates whether $F_T$ and $V_X$ are correlated. As shown by \cite{goodfellow2014generative}, $p(V_X, F_T)=p(V_X)p(F_T)$ when the Nash equilibrium is achieved, thus minimizing the $I(T; V_X)$. The $I(V; T_X)$ can be optimized similarly.
Moreover, note that $I(V_S;V_X)$ and $I(T_S;T_X)$ in $L_{reg}$ can be estimated akin to $I(V_X,F_T)$.

\section{Theoretical Analyses}
\label{sec: theoretical}
\subsection{Cross-Modal Disentanglement}
In this section, we theoretically analyze DisNCL's disentanglement efficacy, pivotal for DisNCL to mitigate the impact of MEI. 
This theoretical exploration is structured into three facets: the completeness, disentanglement, and minimal sufficiency of learned representations. The minimal sufficiency is further divided into two separate components: the minimal sufficiency of modality-invariant information (MII) and modality-exclusive information (MEI).
\begin{theorem}
  Given a multi-modal input pair $(V,T)$ with disentangled representations $F_V=(V_S,V_X)$ and $F_T=(T_S,T_X)$ achieved by optimizing $L_{Dis}$, we have:
  \begin{itemize}[leftmargin=3mm]
    \item \textbf{Completeness}: The optimal representations $F_V^*$ and $F_T^*$ corresponding to multi-modal input $(V,T)$ are sufficient, i.e., 
    $I(F_T^*,T)=H(T)$ and $I(F_V^*,V)=H(V)$.
    \item \textbf{Mutual Disentanglement}: The optimal modality-invariant representations and modality-exclusive representations are disentangled, i.e., $I(V_S^*;V_X^*)=0$ and $I(T_S^*;T_X^*)=0$.
    \item \textbf{Minimal Sufficiency of $V_S$ and $T_S$}: The optimal modality-invariant representations $V_S^*$ and $T_S^*$ are minimal sufficient, i.e., $I(V;V_S^*|T)=0$ and $I(T;T_S^*|V)=0$.
    \item \textbf{Minimal Sufficiency of $V_X$ and $T_X$}: The optimal modality-exclusive representations $V_X^*$ and $T_X^*$ are minimal sufficient, i.e., $I(V;V_X^*)=H(V|T)$ and $I(T;T_X^*)=H(T|V)$.
  \end{itemize}
  \begin{proof}
      Please refer to Appendix A-C for detailed proof.
  \end{proof}
  \label{th: theoretical analyses}
\end{theorem}
Specifically, the completeness ensures that $F_V$ and $F_T$ contains all information of $V$ and $T$ losslessly, while the mutual disentanglement indicates that there is no overlap between $(V_S,V_X)$ and $(T_S,T_X)$.
Moreover, the minimal sufficiency underscores the fidelity of learned representations, \eg, $I(V;V_S^*|T)=0$ indicates that $V_S^*$ adeptly isolates MII, free from any modality-exclusive disturbances. 
Concurrently, $I(V;V_X^*)=H(V|T)$ indicates that $V_X^*$ encodes the sole MEI of $V$, devoid of any modality-invariant perturbations.
Crucially, these theoretical constraints jointly formulate the desired modality-invariant and modality-specific subspace. 
Due to the completeness and minimal sufficiency, $V_S^*$ and $T_S^*$ captures all modality-invariant information, untainted by modality-exclusive elements. Simultaneously, due to mutual disentanglement and minimal sufficiency, $V_X^*$ and $T_X^*$ capture all modality-exclusive information of $V$ and $T$ without any modality-invariant noise.

\noindent\textbf{Discussion}: Previous cross-modal disentanglement methods fall short in noise tolerance and disentanglement efficacy with their trivial MII extraction strategies.
Specifically, they introduce feature consistency \cite{yang2022disentangled,gu2022cross}, cross prediction \cite{wang2024semantics,chen2020rgbd,tian2022multimodal} and information regularization \cite{mao2021deep,tian2021farewell} to ensure the optimal modality-invariant representations. 
Although intuitive, the feature consistency and cross-prediction strategies are sensitive to noisy correspondences and may capture only a subset of the optimal MII with the remainder leaking into MEI.
Moreover, information regularization methods introduce various regularizers on MII to discard redundant input information, thereby showcasing improved noise tolerance. However, it still fails on disentanglement efficacy due to inevitable performance-compression trade-off of trivial regularizers \cite{pan2021disentangled}, \ie, the performance decreases as the compression level intensifies.
In contrast, with appealing noise tolerance shown in \Cref{th: noise robust}, our core theoretical contribution rigorously proves DisNCL's optimal cross-modal disentanglement efficacy, which is unprecedented in previous works and provides a theoretical guarantee for superior efficacy.
We show detailed validation and comparison in \Cref{sec: disentanglement,sec: disentanglement comparsion}.

\subsection{Noise Robust}
This section analyzes the noise robustness of our fine-grained cross-modal alignment strategy. Specifically, given overall noise rate $\eta\in [0,1]$, we follow \cite{hu2023cross} to assume uniform distribution of label noise on $\mathbf{e}_i$, $\eta_{ij} = p(\tilde{y} = j | y = i)=\frac{\eta}{C-1}$.
In this sense, $L_{soft}$ is noise-robust when $\beta_3=1$, \ie
\begin{theorem}
Let $[1,\dots,C]$ be label set where $C$ equals batch size $B$ denoting the label set size, given uniform label noise with $\eta < 1 - \frac{1}{C}$, $L_{soft}$ is noise robust when $\beta_3=1$.
\label{th: noise robust}
\begin{proof}
  Please refer to Appendix A-E for detailed proof.
\end{proof}
\end{theorem}
Moreover, as in Appendix A-E, we follow \cite{ghosh2017robust,feng2021can} to demonstrate the noise tolerance of $L_{soft}$ when $\beta_3\in[0,1]$, attributed to its boundedness.
Collectively, these analyses provably demonstrate the certifiable noise tolerant ability of our fine-grained cross-modal alignment strategy, highlighting DisNCL's appealing efficacy in noisy correspondence learning.

\section{Experiments}
For evaluating our DisNCL, we first introduce details of datasets and implementations. Then we discuss the experimental results. Specifically, we aim to answer following questions:
\begin{itemize}[leftmargin=3mm]
    \item \textbf{RQ1}: Whether  DisNCL can achieve superior performance on varying noisy cross-modal retrieval benchmarks?
    \item \textbf{RQ2}: Does  DisNCL achieves cross-modal disentanglement?
    \item \textbf{RQ3}: How does DisNCL's component facilitate disentanglement?
\end{itemize}

\begin{table*}[!ht]
	\newcommand{\tabincell}[2]{\begin{tabular}{@{}#1@{}}#2\end{tabular}}
	\centering
	\caption{Image-Text Retrieval on Flickr30K and MS-COCO 1K with Synthetic Noise by Randomly Shuffling Images (DECL-Style), where `\dag' signifies methods that incorporate priors such as  additional clean samples or extra model ensemble.}
	\label{table:flicker}
	\resizebox{\textwidth}{!}{ 
		\begin{tabular}{c|c|ccc|ccc|c|ccc|ccc|c}
		\toprule[1.5pt]
		&&\multicolumn{7}{c|}{Flickr30K}&\multicolumn{7}{c}{MS-COCO 1K}\\
		&&\multicolumn{3}{c|}{Image$\longrightarrow$Text}&\multicolumn{3}{c|}{Text$\longrightarrow$Image}&&\multicolumn{3}{c|}{Image$\longrightarrow$Text}&\multicolumn{3}{c|}{Text$\longrightarrow$Image}&\\
		\hline
			Noise&Methods&R@1&R@5&R@10&R@1&R@5&R@10&R\_Sum&R@1&R@5&R@10&R@1&R@5&R@10&R\_Sum\\
			\midrule
			\multirow{9}{*}{20\%}&SCAN& 58.5   & 81.0  & 90.8 & 35.5 & 65.0  & 75.2 & 406.0  & 62.2 & 90.0  & 96.1 & 46.2 & 80.8 & 89.2 & 464.5\\
			~&SAF & 62.8   & 88.7 & 93.9 & 49.7 & 73.6 & 78.0  & 446.7 & 71.5 & 94.0  & 97.5 & 57.8 & 86.4 & 91.9 & 499.1 \\
			~&NCR & 73.5   & 93.2 & 96.6 & 56.9 & 82.4 & 88.5 & 491.1 & 76.6 & 95.6 & 98.2 & 60.8 & 88.8 & 95.0  & 515.0  \\
			~&DECL& 77.5 &93.8 &97.0&56.1&81.8&88.5&494.7&77.5&95.9&98.4&61.7&89.3&95.4&518.2\\
			~&BiCro\dag&74.7&94.3&96.8&56.6&81.4&88.2&492.0&76.6&95.4&98.2&61.3&88.8&94.8&515.1\\
			~&MSCN\dag&75.0&94.5&96.8&56.8&81.4&88.1&492.6&77.0&95.5&98.4&61.6&88.7&94.9&516.1\\
			~&RCL&75.9&94.5&97.3&57.9&82.6&88.6&496.8&78.9&96.0&98.4&62.8&89.9&95.4&521.4\\
			~&CRCL&77.2&93.4&97.3&59.1&83.0&89.6&499.6&79.1&96.1  &98.6  &63.1  &90.3  & 95.7 & 522.9 \\
			~&SREM&77.0&93.7&96.6&57.6&82.5&89.1&496.5&78.4&96.0  &98.5  &62.9  &89.9  & 95.5& 521.2 \\
			~&\textbf{DisNCL}&\textbf{79.1}&\textbf{95.5}&\textbf{98.0}&\textbf{60.3}&\textbf{84.9}&\textbf{90.3}&\textbf{508.1}&\textbf{80.4} & \textbf{96.4} & \textbf{98.9} & \textbf{64.8} & \textbf{90.8}& \textbf{96.1} & \textbf{527.4} \\
			\midrule
			\multirow{9}{*}{40\%}&SCAN& 26.0    & 57.4 & 71.8 & 17.8 & 40.5 & 51.4 & 264.9 & 42.9 & 74.6 & 85.1 & 24.2 & 52.6 & 63.8 & 343.2 \\
			~&SAF & 7.4   & 19.6 & 26.7 & 4.4 & 12.2 & 17.0  & 87.3 & 13.5 & 43.8 & 48.2 & 16.0  & 39.0  & 50.8 & 211.3 \\
			~&NCR & 68.1   & 89.6 & 94.8 & 51.4 & 78.4 & 84.8 & 467.1 & 74.7 & 94.6 & 98.0  & 59.6 & 88.1 & 94.7 & 509.7 \\
			~&DECL&72.7&92.3&95.4&53.4&79.4&86.4&479.6&75.6&95.5&98.3&59.5&88.3&94.8&512.0\\
			~&BiCro\dag&70.7&92.0&95.5&51.9&77.7&85.4&473.2&75.2&95.3&98.1&60.0&87.8&94.3&510.7\\
			~&MSCN\dag&71.9&92.0&95.4&55.1&80.2&86.8&481.3&77.1&95.7&98.4&61.2&88.6&94.8&515.7\\
			~&RCL&72.7&92.7&96.1&54.8&80.0&87.1&483.4&77.0&95.5&98.3&61.2&88.5&94.8&515.3\\
			~&CRCL&72.9&92.7&96.1&55.7&81.0&87.9&486.3& 77.2 & 95.3 & 98.4 & 61.7 & 89.2 & 95.2 & 517.0  \\
			~&SREM&72.8&92.7&96.2&55.6&80.8&87.6&485.7&77.1 &95.6 &98.3 &61.5 &88.7 &94.9 & 516.1 \\
			~&\textbf{DisNCL}&\textbf{76.1}&\textbf{93.2}&\textbf{97.2}&\textbf{58.1}&\textbf{82.5}&\textbf{88.9}&\textbf{496.0}&\textbf{78.5} & \textbf{96.2} & \textbf{98.6} & \textbf{62.8} & \textbf{89.8}& \textbf{95.5} & \textbf{521.4} \\
			\midrule
			\multirow{9}{*}{60\%}&SCAN& 13.6     & 36.5 & 50.3 & 4.8  & 13.6 & 19.8 & 138.6 & 29.9 & 60.9 & 74.8 & 0.9  & 2.4  & 4.1  & 173.0   \\
			~&SAF  & 0.1      & 1.5  & 2.8  & 0.4  & 1.2  & 2.3  & 8.3   & 0.1  & 0.5  & 0.7  & 0.8  & 3.5  & 6.3  & 11.9  \\
			~&NCR  & 13.9     & 37.7 & 50.5 & 11.0   & 30.1 & 41.4 & 184.6 & 0.1  & 0.3  & 0.4  & 0.1  & 0.5  & 1.0    & 2.4   \\
			~&DECL&65.2&88.4&94.0&46.8&74.0&82.2&450.6&73.0&94.2&97.9&57.0&86.6&93.8&502.5\\
			~&BiCro\dag&64.1&87.1&92.7&47.2&74.0&82.3&447.4&73.2&93.9&97.6&57.5&86.3&93.4&501.9\\
			~&MSCN\dag&67.5&88.4&93.1&48.7&76.1&82.3&456.1&74.1&94.4&97.6&57.5&86.4& 93.4& 503.4\\
			~&RCL&67.7&89.1&93.6&48.0&74.9&83.3&456.6&74.0&94.3&97.5&57.6&86.4&93.5&503.3\\
			~&CRCL&67.9&89.2&94.0&48.7&75.3&83.4&458.5&74.2 &94.3 &97.7 &58.1 &87.4 &94.0 & 505.7 \\
			~&SREM&67.9&89.1&93.7&48.6&75.2&83.2&457.7&74.1 &94.3 &97.7 &57.7 &86.8 &93.6 & 504.2 \\
			~&\textbf{DisNCL}&\textbf{68.4}&\textbf{90.3}&\textbf{95.8}&\textbf{51.1}&\textbf{78.2}&\textbf{85.5}&\textbf{469.3}&\textbf{76.5} & \textbf{95.4} & \textbf{98.1} & \textbf{60.6} & \textbf{88.5}& \textbf{94.9} & \textbf{514.1} \\
			
			\bottomrule[1.5pt]
	\end{tabular}}
 \vspace{-4mm}
\end{table*}

\subsection{Experiments Setting}
\subsubsection{Datasets}

For thorough evaluations of our DisNCL, we follow \cite{qin2022deep,huang2021learning} to select three widely-used benchmarks, \ie,
\begin{itemize}
    \item \textbf{Flickr30K} \cite{young2014image} comprises 31,783 images with five captions each from Flickr website. Following \cite{huang2021learning}, we allocate 1K images for validation, another 1K for testing, and the rest for training.
    \item \textbf{MS-COCO} \cite{lin2014microsoft} contains 123,287 images, where each image is associated with five captions. Similar to \cite{huang2021learning}, we use 5K images for validation, another 5K for testing, and the rest for training.
    \item \textbf{CC152K} \cite{huang2021learning} is a subset of Conceptual Captions \cite{sharma2018conceptual}, selected by \cite{huang2021learning}, containing 152K image-text pairs crawled from the Internet. Due to the absence of manual annotation, CC152K naturally contains around 20\% mismatched sample pairs, \ie, real-world noisy correspondences. 
    Following \cite{huang2021learning}, we use 150K images for training, 1K images for validation and another 1K for testing.
\end{itemize}

\subsubsection{Evaluation Metrics}
Following \cite{huang2021learning,han2023noisy,qin2022deep}, we evaluate DisNCL with the standard retrieval metric, R@K.
R@K measures the proportion of queries for which the correct item is retrieved within the top K closest points to the query.
We systematically report the corresponding results of R@1, R@5, and R@10 in both image-to-text and text-to-image retrieval scenarios.
These metrics are further aggregated to evaluate the overall performance, denoted as R\_sum.
\subsubsection{Implementation Details}
As a versatile cross-modal disentanglement framework, DisNCL can be seamlessly integrated into various image-text retrieval methods for enhanced robustness against noisy correspondences during training. 
Here, we employ SGRAF \cite{diao2021similarity} backbone with same settings as \cite{huang2021learning,han2023noisy,qin2022deep} for fair comparisons.
Specifically, we conduct retrieval training and inference within the learned modality-invariant subspace.
To begin with, we employ a `warm-up' phase using $L_{align}$ with constant margin $\alpha=0.2$ for 5 epochs to achieve initial convergence. 
Then, the model is trained for 50 epochs using $L_{DisNCL}$ with an Adam optimizer, whose initial learning rate is 2e-4 and is reduced to 0.1x after 25 epochs.
Following \cite{han2023noisy,qin2022deep}, all experiments adopt shared hyperparameters: $\tau=0.05$,  $m=10$, a batch size of 128, a word embedding size of 300, and a unified feature space dimension of 1,024, \etc. 
For cross-modal disentanglement, we set the dimensions of \( V_S, V_X, T_S, T_X \) to 512. 
The hyperparameter \( \gamma \) is set to 0.5; \( \beta_1, \beta_2 \) are set to 0.1 while $\beta_3$ is set to 0.5.
The discriminators and disentangled encoders, \eg, \( f_1 \), \( f_S \) and $g_S$, are implemented as three-layer multi-layer perceptrons with LeakyReLU activation (\( \alpha = 0.2 \)) and 256 hidden dimension.

\subsection{Comparison with State-Of-The-Art (SOTA)}
To answer \textbf{RQ1}, we compare DisNCL against current SOTA cross-modal retrieval methods to demonstrate its efficacy, including two main categories of baselines: traditional methods with well-matched data assumption, \eg, SCAN \cite{lee2018stacked} and SAF \cite{diao2021similarity}; and methods robust to noisy correspondence, including NCR \cite{huang2021learning}, DECL \cite{qin2022deep}, MSCN \cite{han2023noisy}, BiCro \cite{yang2023bicro}, RCL \cite{yang2021partially}, CRCL \cite{qin2023cross}, SREM \cite{dang2024noisy}, ESC \cite{yang2024robust}, GSC \cite{zhao2024mitigating} and CREAM \cite{ma2024cross}.
Moreover, as Flickr30K and MS-COCO are well-matched datasets annotated by human, we carry out experiments by generating the synthesized mismatched pairs.
Specifically, we follow NCR \cite{huang2021learning} and DECL \cite{qin2022deep} to conduct experiments with two noise generation methods for comprehensive evaluation of our DisNCL. The detailed noise generation methods are summarized as follows:
\begin{itemize}
  \item DECL \cite{qin2022deep} and RCL \cite{hu2023cross} inject noisy correspondence in Flickr30K and MS-COCO datasets by \textbf{randomly shuffling images} for a specific percentage (noise ratio). 
  \item  NCR \cite{huang2021learning} and MSCN \cite{han2023noisy} randomly select a specific percentage (noise ratio) of images and \textbf{randomly permute all their corresponding captions}.
\end{itemize}

\begin{table*}[!ht]
	\newcommand{\tabincell}[2]{\begin{tabular}{@{}#1@{}}#2\end{tabular}}
	\centering
	\caption{Image-Text Retrieval on Flickr30K and MS-COCO 1K with Synthetic Noise by Randomly Shuffling Captions (NCR-Style), where `\dag' signifies methods that incorporate priors such as  additional clean samples or extra model ensemble.}
	\label{table:flicker_NCR}
	\resizebox{\textwidth}{!}{ 
		\begin{tabular}{c|c|ccc|ccc|c|ccc|ccc|c}
		\toprule[1.5pt]
		&&\multicolumn{7}{c|}{Flickr30K}&\multicolumn{7}{c}{MS-COCO 1K}\\
		&&\multicolumn{3}{c|}{Image$\longrightarrow$Text}&\multicolumn{3}{c|}{Text$\longrightarrow$Image}&&\multicolumn{3}{c|}{Image$\longrightarrow$Text}&\multicolumn{3}{c|}{Text$\longrightarrow$Image}&\\
		\hline
			Noise&Methods&R@1&R@5&R@10&R@1&R@5&R@10&R\_Sum&R@1&R@5&R@10&R@1&R@5&R@10&R\_Sum\\
			\midrule
			\multirow{9}{*}{20\%}&SCAN& 58.5   & 81.0  & 90.8 & 35.5 & 65.0  & 75.2 & 406.0  & 62.2 & 90.0  & 96.1 & 46.2 & 80.8 & 89.2 & 464.5\\
			~&NCR & 73.5   & 93.2 & 96.6 & 56.9 & 82.4 & 88.5 & 491.1 & 76.6 & 95.6 & 98.2 & 60.8 & 88.8 & 95.0  & 515.0  \\
			~&BiCro\dag&78.1&94.4&97.5&60.4&84.4&89.9&504.7&78.8&96.1&98.6&63.7&90.3&95.7&523.2\\
			~&MSCN\dag&77.4&94.9&97.6&59.6&83.2&89.2&502.1&78.1&97.2&98.8&64.3&90.4&95.8&524.6\\
			~&RCL&77.5&94.6&97.0&59.5&83.9&89.8&502.3&79.4&96.3&98.8&63.8&90.3&95.5&524.1\\
			~&ESC\dag&79.0&94.8&97.5&59.1&83.8&89.1&503.3&79.2&97.0 &99.1  &64.8  &90.7  & 96.0 & 526.8 \\
			~&GSC\dag&78.3&94.6&97.8&60.1&84.5&90.5&505.8&79.5&96.4 &98.9  &64.4  &90.6  & 95.9 & 525.7 \\
			~&CREAM&77.4&95.0&97.3&58.7&84.1&89.8&502.3&78.9&96.3 &98.6  &63.3  &90.1  & 95.8 & 523.0 \\
			~&\textbf{DisNCL}&\textbf{79.3}&\textbf{95.0}&\textbf{98.0}&\textbf{60.6}&\textbf{85.0}&\textbf{90.5}&\textbf{508.4}&\textbf{80.6} & \textbf{96.6} & \textbf{98.6} & \textbf{64.6} & \textbf{90.7}& \textbf{96.0} & \textbf{527.2} \\
			\midrule
			\multirow{9}{*}{40\%}&SCAN& 26.0    & 57.4 & 71.8 & 17.8 & 40.5 & 51.4 & 264.9 & 42.9 & 74.6 & 85.1 & 24.2 & 52.6 & 63.8 & 343.2 \\
			~&NCR & 68.1   & 89.6 & 94.8 & 51.4 & 78.4 & 84.8 & 467.1 & 74.7 & 94.6 & 98.0  & 59.6 & 88.1 & 94.7 & 509.7 \\
			~&BiCro\dag&67.6&90.8&94.4&51.2&77.6&84.7&466.3&73.9&94.4&97.8&58.3&87.2&93.9&505.5\\
			~&MSCN\dag&67.5&88.4&93.1&48.7&76.1&82.3&456.1&74.1&94.4&97.6&57.5&86.4& 93.4& 503.4\\
			~&RCL&76.2&92.5&96.2&55.9&80.9&87.5&489.2&78.1&95.4&98.2&62.5&89.0& 95.2& 518.4\\
			~&ESC\dag&76.1&93.1&96.4&56.0&80.8&87.2&489.6&78.6& \textbf{96.6} & \textbf{99.0}  &63.2  & \textbf{90.6}  & \textbf{95.9} & \textbf{523.9} \\
			~&GSC\dag&76.5&94.1&\textbf{97.6}&57.5&82.7&88.9&497.3&78.2&95.9 &98.2  &62.5  &89.7  & 95.4 & 519.9 \\
			~&CREAM&76.3&93.4&97.1&57.0&82.6&88.7&495.1&76.5&95.6 &98.3  &61.7  &89.4  & 95.3 & 516.8 \\
			~&\textbf{DisNCL}&\textbf{77.1}&\textbf{94.1}&97.1&\textbf{58.7}&\textbf{82.9}&\textbf{89.5}&\textbf{499.4}&\textbf{79.0} & 96.2 & 98.6 & \textbf{63.5} & 90.1 & 95.6 & 523.0 \\
			\midrule
			\multirow{9}{*}{60\%}&SCAN& 13.6     & 36.5 & 50.3 & 4.8  & 13.6 & 19.8 & 138.6 & 29.9 & 60.9 & 74.8 & 0.9  & 2.4  & 4.1  & 173.0   \\
			~&NCR  & 13.9     & 37.7 & 50.5 & 11.0   & 30.1 & 41.4 & 184.6 & 0.1  & 0.3  & 0.4  & 0.1  & 0.5  & 1.0    & 2.4   \\
			~&BiCro\dag&67.6&90.8&94.4&51.2&77.6&84.7&466.3&73.9&94.4&97.8&58.3&87.2&93.9&505.5\\
			~&MSCN\dag&67.5&88.4&93.1&48.7&76.1&82.3&456.1&74.1&94.4&97.6&57.5&86.4& 93.4& 503.4\\

			~&RCL&71.8&90.6&94.1&52.9&78.7&85.2&473.3&75.3&94.8 &97.8  &60.4  &88.1  & 94.4 & 510.8 \\
			~&ESC\dag&72.6&90.9&94.6&53.0&78.6&85.3&475.0&77.2&95.1 &98.1  &61.1  &88.6  & 94.9 & 515.0 \\
			~&GSC\dag&70.8&91.1&95.9&53.6&79.8&86.8&478.0&75.6&95.1 &98.0  &60.0  &88.3  & 94.6 & 511.7 \\
			~&CREAM&70.6&91.2&96.1&53.3&79.2&87.0&477.4&74.7&94.8 &98.0  &59.7  &88.0  & 94.6 & 509.9 \\

			~&\textbf{DisNCL}&\textbf{72.5}&\textbf{92.5}&\textbf{95.9}&\textbf{54.2}&\textbf{80.2}&\textbf{87.0}&\textbf{482.3}&\textbf{77.2} & \textbf{95.5} & \textbf{98.1} & \textbf{61.5} & \textbf{88.7}& \textbf{95.1} & \textbf{516.1} \\
			
			\bottomrule[1.5pt]
	\end{tabular}}
 \vspace{-4mm}
\end{table*}

\subsubsection{\textbf{Synthetic Noise}}
As aforementioned, we inject noise into well-matched Flickr30K and MS-COCO datasets and report results with noise ratio 20\%, 40\%, 60\% for comprehensive comparison with current SOTAs in \Cref{table:flicker,table:flicker_NCR}. 

Following \cite{qin2023cross,han2023noisy}, the MS-COCO results are computed by averaging over 5 folds of 1K test images.
We can observe that DisNCL consistently achieves the best R@1 performance across all noise ratios and types.
Notably, with noise ratio increasing, traditional methods SCAN \cite{lee2018stacked} and SAF \cite{diao2021similarity} fail, while noise-robust methods, \eg, NCR \cite{huang2021learning} and DECL \cite{qin2022deep},  are less sensitive with less performance drop, revealing the necessity for models adept at effectively withstanding noise.

However, although effective, these noise-robust methods' performance drops sharply with noise ratios increasing.
This decline can be attributed to the growing sensitivity of these methods to sub-optimal similarity predictions within entangled feature spaces, especially under high noise ratios. 
Specifically, the sub-optimal similarity predictions hinder the identification of noisy correspondences and fail to suppress their negative impact during training.
Consequently, such limitations exacerbate the propagation of incorrect supervisory signals during the training process. 
Conversely, owing to its optimal cross-modal disentanglement, DisNCL preserves more accurate similarity predictions and soft matching probabilities in modality-invariant subspace, consistently outperforming strong competitors by 2\% on average. Crucially, this performance advantage is more pronounced as the noise ratio increases. These findings highlight DisNCL's efficacy and adaptability to noisy correspondences, underscoring its appealing applicability.

\begin{table}[t!]
	\caption{Image-Text Retrieval on Real-World Noise CC152K.}
	\makeatletter\def\@captype{table}
	\resizebox{\linewidth}{!}{
		\begin{tabular}{c|ccc|ccc|c}
			\toprule[1.5pt]
			& \multicolumn{3}{c|}{Image$\longrightarrow$Text}&\multicolumn{3}{c|}{Text$\longrightarrow$Image}&\\
			\hline
			Methods&R@1&R@5&R@10&R@1&R@5&R@10&R\_Sum\\
			\hline
			SCAN&30.5&55.3&65.3&26.9&53.0&64.7&295.7\\
			SAF&31.7&59.3&68.2&31.9&59.0&67.9&318.0\\
			NCR&39.5&64.5&73.5&40.3&64.6&73.2&355.6\\
			DECL&39.0&66.1&75.5&40.7&66.3&76.7&364.3\\
			BiCro\dag&40.8&67.2&76.1&42.1&67.6&76.4&370.2\\
			RCL&41.7&66.0&73.6&41.6&66.4&75.1&364.4\\
			MSCN\dag&40.1&65.7&76.6&40.6&67.4&76.3&366.7\\
			CRCL&41.8&67.4&76.5&41.6&68.0&\textbf{78.4}&373.7\\
			CREAM&40.3&68.5&77.1&40.2&68.2&78.3&372.6\\
			GSC\dag&42.1&68.4&\textbf{77.7}&42.2&67.6&77.1&375.1\\
			ESC\dag&42.8&67.3&76.9&\textbf{44.8}&68.2&75.9&375.9\\
			\textbf{Ours}&\textbf{42.9}&\textbf{68.5}& 76.9 &43.9&\textbf{69.4}&77.6&\textbf{379.1}\\
			
			\bottomrule[1.5pt]
		\end{tabular}
	}
  \label{cc152k}
  \vspace{-4mm}
  \end{table}

  \begin{figure*}[!t]
    \includegraphics[width=1\textwidth]{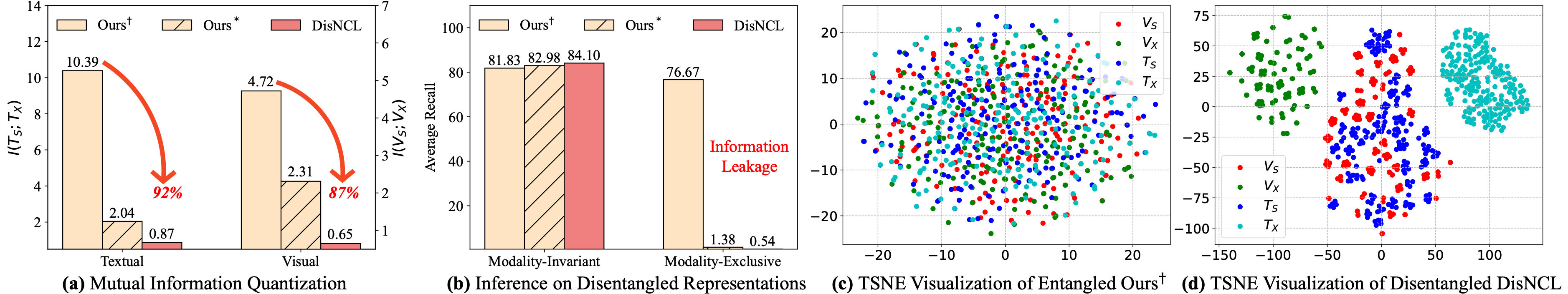}
    \caption{Ablation on disentanglement analysis, where Ours$^\dag$ and Ours$^*$ refer to DisNCL w/o $L_{Dis}+L_{reg}$ and $L_{reg}$. Ours$^\dag$ is further trained with using $(V_S,V_X), (T_S,T_X)$ to reconstruct $(V,T)$, ensuring disentangled representations capture all input information.}
    \label{fig:disentanglement}
    \vspace{-4mm}
  \end{figure*}

\subsubsection{\textbf{Real-World Noise}}

The web-crawled CC152K inherently comprises around 20\% noisy correspondences \cite{huang2021learning}, offering a more realistic noise setting.
Therefore, we directly conduct experiments on CC152K to evaluate DisNCL in real-world noise setting without any additional noise injection.
\Cref{cc152k} shows that DisNCL consistently improves performance over all baselines, showcasing its superior real-world robustness.
Notably, DisNCL surpasses MSCN, which utilizes additional clean data, and GSC, which employs extra model ensemble, by  1\% in R\_sum, highlighting DisNCL's remarkable adaptability and efficacy in real-world settings.

\begin{figure*}[!t]
  \includegraphics[width=\textwidth]{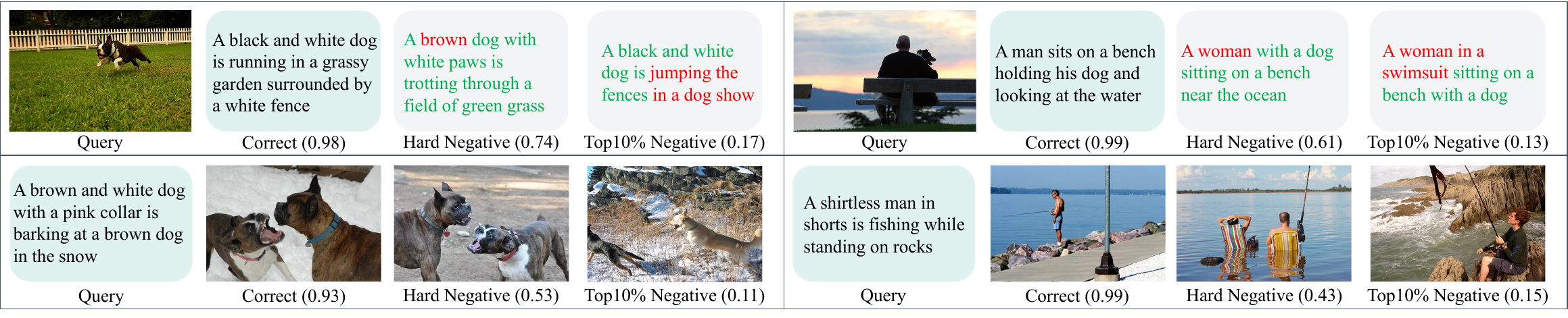}
  \caption{Soft target visualization in both image-to-text (above) and text-to-image (below) retrieval on 20\% noise Flickr30K.}
  \label{fig:soft_visualization}
  \vspace{-4mm}
\end{figure*}

\begin{table}[t!]
  \centering
  \caption{Comparison with CLIP and previous SOTAs on MS-COCO 5K, where CLIP* denote the original pretrained CLIP's zero-shot inference. In 0\% noise, all methods are directly trained/fine-tuned on original MS-COCO with no noise.}
    \label{table: CLIP comparison}
    \makeatletter\def\@captype{table}
      \begin{tabular}{c|c|c|c|c}
        \toprule[1.5pt]
        \multirow{2}{*}{Noise}& \multirow{2}{*}{Methods}& \multicolumn{1}{c|}{Image$\longrightarrow$Text}&\multicolumn{1}{c|}{Text$\longrightarrow$Image}&\multirow{2}{*}{R\_sum}\\
        \cline{3-4}
         & &R@1/5/10&R@1/5/10\\
        \midrule[0.5pt]
        \multirow{6}{*}{0\%}&CLIP*&50.2/74.6/83.6 &30.4/56.0/66.8 & 361.6 \\
        &CLIP&56.4/82.5/89.9 &40.7/69.4/79.6 & 418.5 \\
        &SGRAF&58.8/84.8/92.1 &41.6/70.9/81.5 &429.4\\
        &DECL&59.2/84.5/91.5 &41.7/70.6/81.1 &428.6 \\
        &SREM&59.4/84.6/91.7 &41.9/71.0/81.3 &429.9\\
        &Ours&61.4/86.5/92.9 &44.3/73.1/83.3 &441.4 \\
        \midrule
        \multirow{2}{*}{20\%}&CLIP&21.4/49.6/63.3 &14.8/37.6/49.6 & 236.3 \\
        &Ours&59.8/85.3/92.2&42.7/72.0/82.2&434.2 \\
        \midrule
        \multirow{2}{*}{40\%}&CLIP& 14.9/32.4/42.3 &9.7/22.3/30.2 & 151.8 \\
        &Ours&58.1/84.3/91.8 &41.0/70.1/81.0 &426.3 \\
        \bottomrule[1.5pt]
      \end{tabular}
    \vspace{-4mm}
  \end{table}

\subsubsection{\textbf{Comparison With Pretrained Model}}
In this section, we compare our DisNCL with large pretrained model CLIP \cite{radford2021learning}. 
Briefly, CLIP is trained on massive web-crawled datasets inevitably involving numerous noisy correspondences. 
This comparison clarifies the comparative efficacy of the big data-based model (CLIP) \vs our DisNCL in addressing the noisy correspondence issue.
Specifically, CLIP posits that using millions of data can ignore the potential noisy correspondence. Conversely, we believe that a meticulously designed algorithm is critical to handling noisy correspondences. 

In detail, following \cite{huang2021learning,qin2023cross}, we initially fine-tune the pre-trained CLIP for 32 epochs using MS-COCO's noisy training data.
Then we evaluate the fine-tuned CLIP and our DisNCL on whole MS-COCO 5K for comprehensive comparisons.
Notably, although CLIP employs millions of image-text pairs for pre-training, its performance drops sharply with noise ratio increasing during finetuning. 
In contrast, our DisNCL showcases much more superior and robust retrieval performance with noisy correspondence, highlighting the necessity of algorithm design and confirming our DisNCL's efficacy.

\subsubsection{\textbf{Well-Matched Data}}
This section further evaluates DisNCL in general image-text retrieval without additional noise. 
Notably, \Cref{table: CLIP comparison} shows that DisNCL surpasses its predecessors by 3\%, despite its primary focus on robustness against noisy correspondences.
We attribute this notable improvement to DisNCL's optimal cross-modal disentanglement, effectively reducing irrelevant associations' impact, thereby sharpening model's focus on information shared across modalities.

\subsection{Disentanglement Analysis}
\label{sec: disentanglement}
For \textbf{RQ2}, after training on 20\% noise Flickr30K, we use MINE \cite{belghazi2018mutual} estimator on clean Flickr30K test set to evaluate DisNCL's disentanglement efficacy.
\Cref{fig:disentanglement} shows that DisNCL compresses mutual information between disentangled representations by around 90\% in both modalities, confirming its disentanglement efficacy.
Subsequent retrieval experiments with these disentangled representations further verified their fidelity. 
Specifically,  entangled Ours$^\dag$ achieves commendable retrieval performance with non-inter-modally shared $(V_X,T_X)$, indicating a severe leakage of $(V_S,T_S)$ into $(V_X,T_X)$. 
In contrast,  DisNCL effectively curtails this leakage,  whose retrieval performance on $(V_X,T_X)$ is akin to random guessing, corroborating DisNCL’s disentanglement efficacy.
Furthermore, a t-SNE visualization on learned representations qualitatively affirms disentanglement efficacy. Note that representations from entangled Ours$^\dag$  are intermixed, confirming the aforementioned information leakage. Conversely, DisNCL’s representations display clear boundaries, forming three distinct clusters corresponding to the modality-invariant, visual-exclusive, and textual-exclusive subspace.
Collectively, these results highlight DisNCL's superior disentanglement efficacy, validating our theoretical analyses.

\subsection{Ablation Study}
\subsubsection{Component Analysis}
\label{sec: component}
To answer \textbf{RQ3}, \Cref{table:ablation} shows ablation studies to evaluate the efficacy of DisNCL's component.
Specifically, within \( L_{DisNCL} \), we set aside the conventional retrieval constraint, \( L_{align} \), and ablate the remainings: \( L_{Dis} \), \( L_{soft} \), and \( L_{reg} \).
Notably, DisNCL with only \( L_{align} \) reduces to a conventional method with similarity-based alleviation strategy for noisy correspondences, severely suffering from misled similarity predictions in entangled feature space. 
Using modality-invariant space, $L_{Dis}$ enables the model to exclude adverse effects of modality-exclusive noise, boosting performance with more accurate similarity predictions for identification and suppression of noisy correspondences.
Moreover, \( L_{soft} \) estimates soft targets within the modality-invariant space, allowing the model to recognize more nuanced many-to-many cross-modal relationships for performance improvements. Lastly, \( L_{reg} \) enhances performance by further advancing disentanglement.
The optimal performance is attained with all components, underscoring the essential role of each in cross-modal disentanglement.

\begin{table}[t!]
  \centering
  \caption{Ablation studies on CC152K with real-world noise.}
    \label{table:ablation}
    \makeatletter\def\@captype{table}
      \begin{tabular}{ccc|c|c|c}
        \toprule[1.5pt]
        \multicolumn{3}{c|}{Methods}& \multicolumn{1}{c|}{Image$\longrightarrow$Text}&\multicolumn{1}{c|}{Text$\longrightarrow$Image}&\multirow{2}{*}{R\_sum}\\
        \cline{1-5}
        \thead{$L_{Dis}$}&\thead{$L_{soft}$}&\thead{$L_{reg}$}&R@1/5/10&R@1/5/10\\
        \midrule[0.5pt]
        & & &39.9/65.9/75.0&40.5/65.8/76.0&363.1\\
        \checkmark&& &40.9/67.8/75.8&41.6/68.0/76.2&370.3\\
        \checkmark&\checkmark& &42.4/68.4/76.8&43.4/68.8/76.4&376.2\\
        \checkmark&\checkmark&\checkmark&\textbf{42.9}/\textbf{68.5}/\textbf{76.9}&\textbf{43.9}/\textbf{69.4}/\textbf{77.6}&\textbf{379.1}\\
        \bottomrule[1.5pt]
      \end{tabular}
    \vspace{-4mm}
  \end{table}

\subsubsection{Softened Target Visualization}

\Cref{fig:soft_visualization} shows similarity predictions in image-text retrieval on 20\% noise Flickr30K dataset, including given queries and their correct samples, hard negatives, and top 10\% negatives.
The similarity scores are proportional to the soft matching probabilities as per \Cref{eq: soft logits}, offering a more intuitive illustration of the estimated soft targets.
Specifically, these results reveal the necessity of modeling noisy many-to-many, rather than one-to-one, correspondences.
Moreover, owing to its optimally disentangled modality-invariant subspace, DisNCL accurately captures local similarities in images and texts to generate more accurate soft targets. 
For instance, in the top-right image-to-text retrieval example, DisNCL successfully discerns the textual misidentification of `a woman' versus the actual `a man', effectively reducing the matching probability by 30\%.
Moreover, as erroneous information increases, \eg, `A woman in a swimsuit,' and correct information diminishes, \eg, the absence of `near the ocean', DisNCL further reduces the matching probability by 90\%. 
These outcomes compellingly demonstrate DisNCL's capacity for more nuanced cross-modal alignment by accurately modeling the complex cross-modal relationships, thereby enhancing overall performance.

\subsubsection{Comparison With Previous Cross-Modal Disentanglement Methods}
\label{sec: disentanglement comparsion}
\begin{figure}[!t]
  \centering
  
  \subfloat[Image-to-Text Retrieval.]{\includegraphics[width=0.45\linewidth]{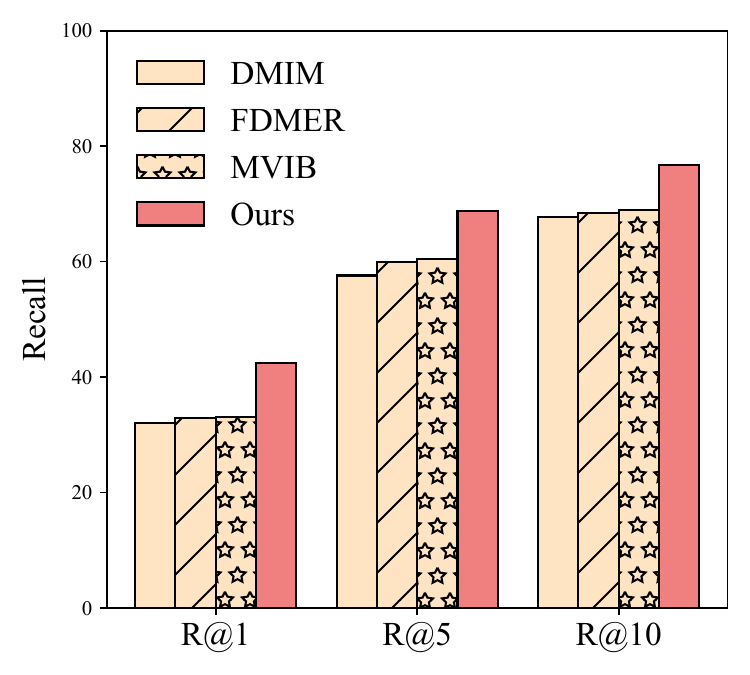}}
  \hspace{3mm}
  \subfloat[Text-to-Image Retrieval.]{\includegraphics[width=0.45\linewidth]{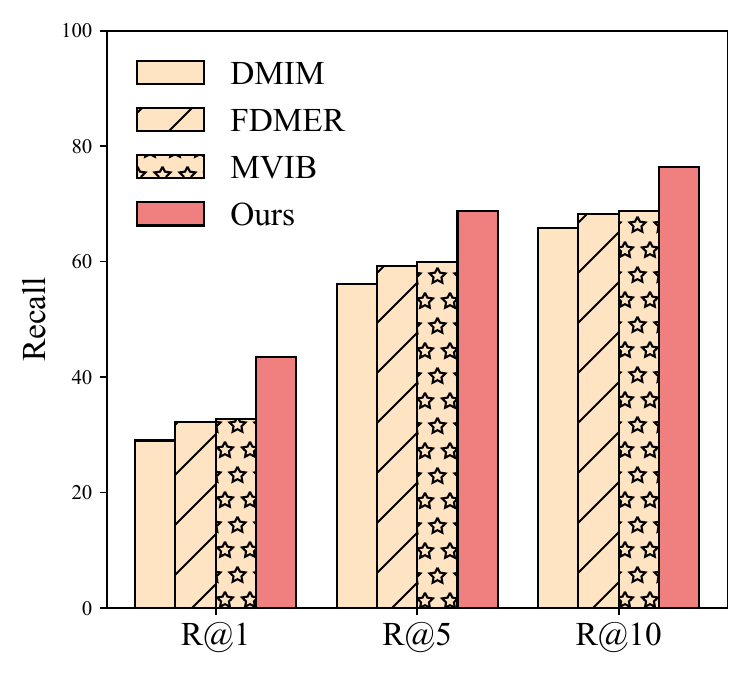}}

	\caption{Comparison with previous representative cross-modal disentanglement methods on real-world noise CC152K.}
  \label{table: FDMER comparison}
  \vspace{-4mm}
\end{figure}

To intuitively illustrate the limitations of previous cross-modal disentanglement methods \cite{gu2022cross,guo2019learning,xia2023achieving,tian2022multimodal,yang2022disentangled} in noisy correspondence scenarios, we conduct experiments by substituting the representative cross-modal disentanglement method, MVIB \cite{federici2020learning} (information regularization), FDMER \cite{yang2022disentangled} (consistency constraint) and DMIM \cite{guo2019learning} (cross prediction), for our DisNCL.
Specifically, FDMER \cite{yang2022disentangled} reduces the squared Frobenius norm differences of modality-invariant representations between modalities, while MVIB \cite{federici2020learning} proposes an information bottleneck to retain modality-invariant information only.
Additionally, DMIM \cite{guo2019learning} exchanges modality-invariant representations during feature reconstruction to ensure the learned representations capture shared information across modalities.
However, these methods cannot ensure the optimal modality-invariant representations, \eg, they may converge trivially to a subset of optimal MII while leaking the remainder to MEI. 
Conversely, with optimal cross-modal disentanglement efficacy and appealing noise tolerance, \Cref{table: FDMER comparison} shows that our DisNCL significantly outperforms these cross-modal disentanglement methods, demonstrating the superior robustness of our DisNCL in noisy correspondence scenarios.
Notably, these results are consistent with our analysis in \Cref{sec: disentanglement}, underscoring the efficacy of DisNCL and validating our theoretical analysis.

\section{Conclusion}
This paper proposes DisNCL, a novel cross-modal disentanglement framework to tackle noisy correspondence prevalent in real-world multi-modal data.
In detail, we introduce an information theoretic objective to naturally decompose multi-modal inputs into complementary modality-invariant and modality-exclusive components with certifiable optimal cross-modal disentanglement efficacy.
Then, we predict similarities within the modality-invariant subspace to exclude adverse effects of modality-exclusive noise, paving the way for more effective identification and suppression of noisy correspondences.
Moreover, we estimate soft targets to model the noisy many-to-many relationship inherent in multi-modal input, thereby facilitating more accurate cross-modal alignment.
Extensive experiments on various benchmarks demonstrate DisNCL's superior efficacy and validate our theoretical analyses.
We hope our DisNCL could provide insights for developing retrieval systems more suitable for real-world scenarios, and inspire future work on cross-modal disentanglement.

{\textbf{Broader Impacts}} Our DisNCL can widely impact various applications that require robust multi-modal understanding and aligning, \eg, multimedia retrieval, image/video caption and recommendation systems \etc 
Specifically, addressing the noisy correspondence issue offers numerous benefits, \eg, significantly reducing the expensive manual data annotation cost; effectively leveraging Internet data despite potential mismatched ones; enhancing multi-modal systems more suitable for noisy real-world scenarios, \etc

\bibliographystyle{IEEEtran}
\bibliography{mybib}
\vspace{-8mm}
\begin{IEEEbiography}[{\includegraphics[width=1in,height=1.25in,clip,keepaspectratio]{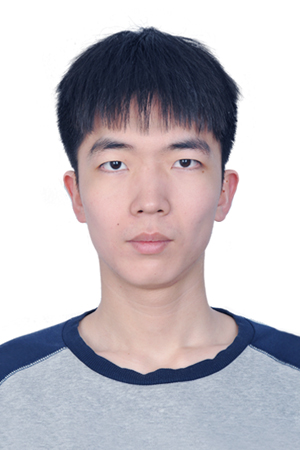}}]{Zhuohang Dang} received a BS degree from the Department of Computer Science and Technology, Xi'an Jiaotong University, in 2021. Currently, he is a Ph.D. student in the Department of Computer Science and Technology at Xi'an Jiaotong University, supervised by Professor Minnan Luo. His research interests include causal inference and computer vision.
\end{IEEEbiography}
\vspace{-8mm}

\begin{IEEEbiography}[{\includegraphics[width=1in,height=1.25in,clip,keepaspectratio]{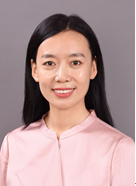}}]{Minnan Luo} received the Ph. D. degree from
the Department of Computer Science and Technology, Tsinghua University, China, in 2014.
Currently, she is a Professor in the
School of Electronic and Information Engineering at Xi’an Jiaotong University. She was a PostDoctoral Research with the School of Computer
Science, Carnegie Mellon University, Pittsburgh,
PA, USA. 
Her research focus in this period was mainly on developing machine learning algorithms and applying them to
computer vision and social networks analysis.
\end{IEEEbiography}
\vspace{-8mm}

\begin{IEEEbiography}[{\includegraphics[width=1in,height=1.25in,clip,keepaspectratio]{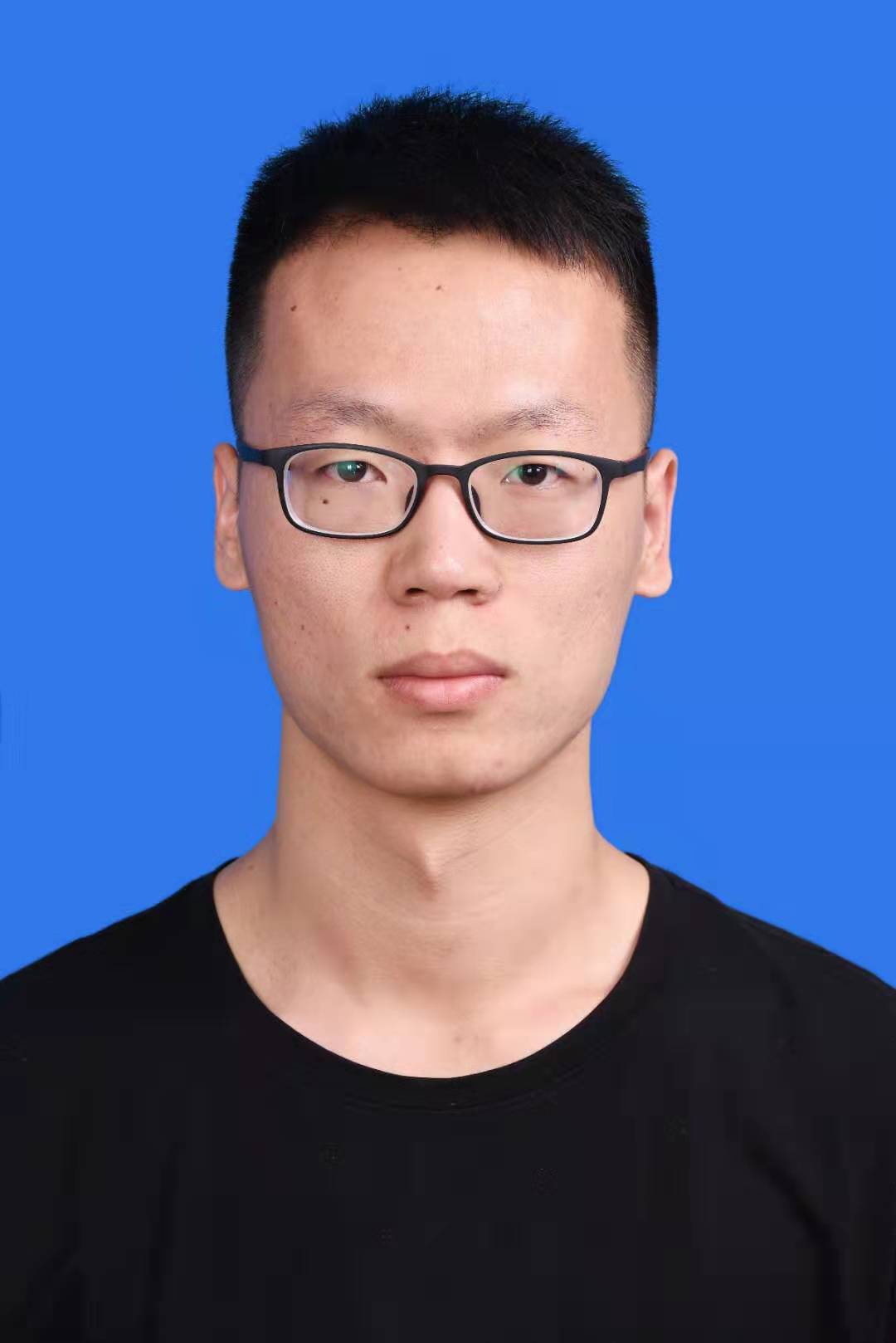}}]{Jihong Wang}
  received the B.Eng. degree from the School of Computer Science and Technology, Xi'an Jiaotong University, China, in 2019. He is currently a Ph.D. student in the School of Computer Science and Technology, Xi'an Jiaotong University, China. His research interests include robust machine learning and its applications, such as social computing and learning algorithms on graphs.
\end{IEEEbiography}
\vspace{-8mm}
\begin{IEEEbiography}[{\includegraphics[width=1in,height=1.25in,clip,keepaspectratio]{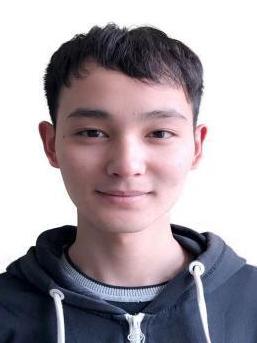}}]{Chengyou Jia} received the BS degree in Computer Science and technology from Xi’an Jiaotong University in 2021. He is currently working
toward the Ph.D. degree in computer science and technology at Xi’an Jiaotong University. His
research interests include machine learning and
optimization, computer vision and multi-modal learning.
\end{IEEEbiography}
\vspace{-8mm}
\begin{IEEEbiography}[{\includegraphics[width=1in,height=1.25in,clip,keepaspectratio]{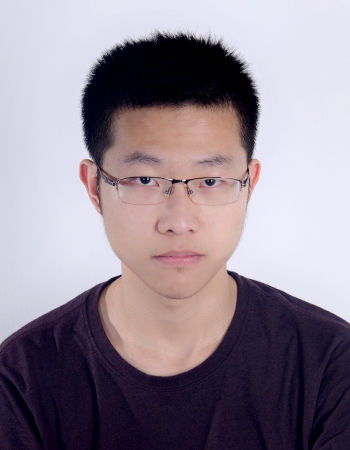}}]{Haochen Han} received the B.S. degree from the School of Energy and Power Engineering, Chongqing University, in 2019. Currently, he is a Ph.D. student in the Department of Computer Science and Technology at Xi'an Jiaotong University, supervised by Professor Qinghua Zheng. His research interests include robust multi-modal learning and its applications, such as action recognition and cross-modal retrieval.
\end{IEEEbiography}
\vspace{-8mm}
\begin{IEEEbiography}[{\includegraphics[width=1in,height=1.25in,clip,keepaspectratio]{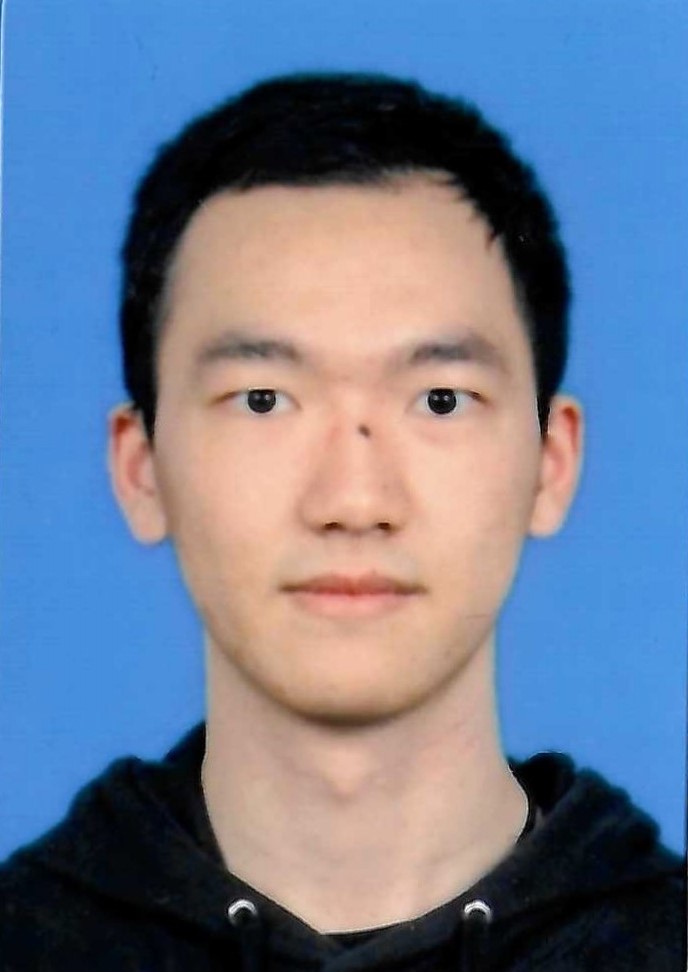}}]{Herun Wan} received a B.S. degree from the Department of Computer Science and Technology, Xi'an Jiaotong University, in 2022. Currently, he is a Ph.D. candidate in the Department of Computer Science and Technology at Xi'an Jiaotong University, supervised by Professor Minnan Luo. His research interests include social network analysis and computer vision.
\end{IEEEbiography}
\vspace{-8mm}

\begin{IEEEbiography}[{\includegraphics[width=1in,height=1.25in,clip,keepaspectratio]{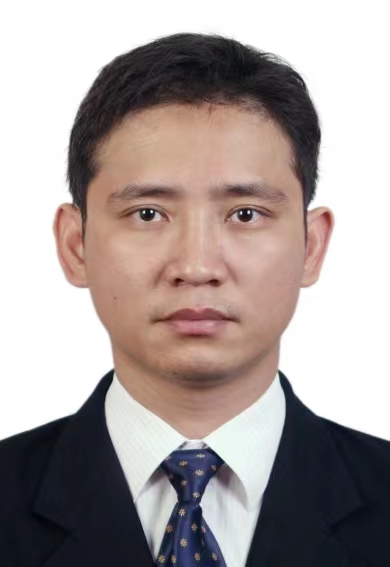}}]{Guang Dai} 
 received his B.Eng. degree in Mechanical Engineering from Dalian University of Technology and M.Phil. degree in Computer Science from the Zhejiang University and the Hong Kong University of Science and Technology. He is currently a senior research scientist at State Grid Corporation of China, and also the founder of SGIT AI Lab,  State Grid Corporation of China. He has published a number of papers at prestigious journals and conferences, e.g., JMLR, AIJ, NeurIPS and ICML. His main research interests include deep learning, reinforcement learning, optimization computation, and related applications.
\end{IEEEbiography}
\vspace{-8mm}
\begin{IEEEbiography}[{\includegraphics[width=1in,height=1.25in,clip,keepaspectratio]{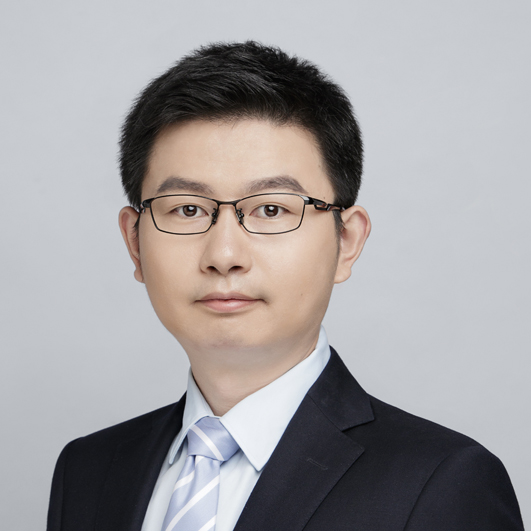}}]{Xiaojun Chang} is a Professor at the School of Information Science and Technology, University of Science and Technology of China. He is also a visiting Professor at Department of Computer Vision, Mohamed bin Zayed University of Artificial Intelligence (MBZUAI). He was an ARC Discovery Early Career Researcher Award (DECRA)
Fellow between 2019-2021. After graduation, he
was worked as a Postdoc Research Associate
in School of Computer Science, Carnegie Mellon University, a Senior Lecturer in Faculty of
Information Technology, Monash University, an Associate Professor in School of Computing Technologies, RMIT University, and a Professor in Faculty of Engineering and Information Technology, University of Technology Sydney. He mainly
worked on exploring multiple signals for automatic content analysis in unconstrained or surveillance videos and has achieved top performance in various international competitions. He received his Ph.D. degree from University of Technology Sydney. His research focus was mainly on developing machine learning algorithms and applying them to
multimedia analysis and computer vision.
\end{IEEEbiography}
\vspace{-8mm}
\begin{IEEEbiography}[{\includegraphics[width=1in,height=1.25in,clip,keepaspectratio]{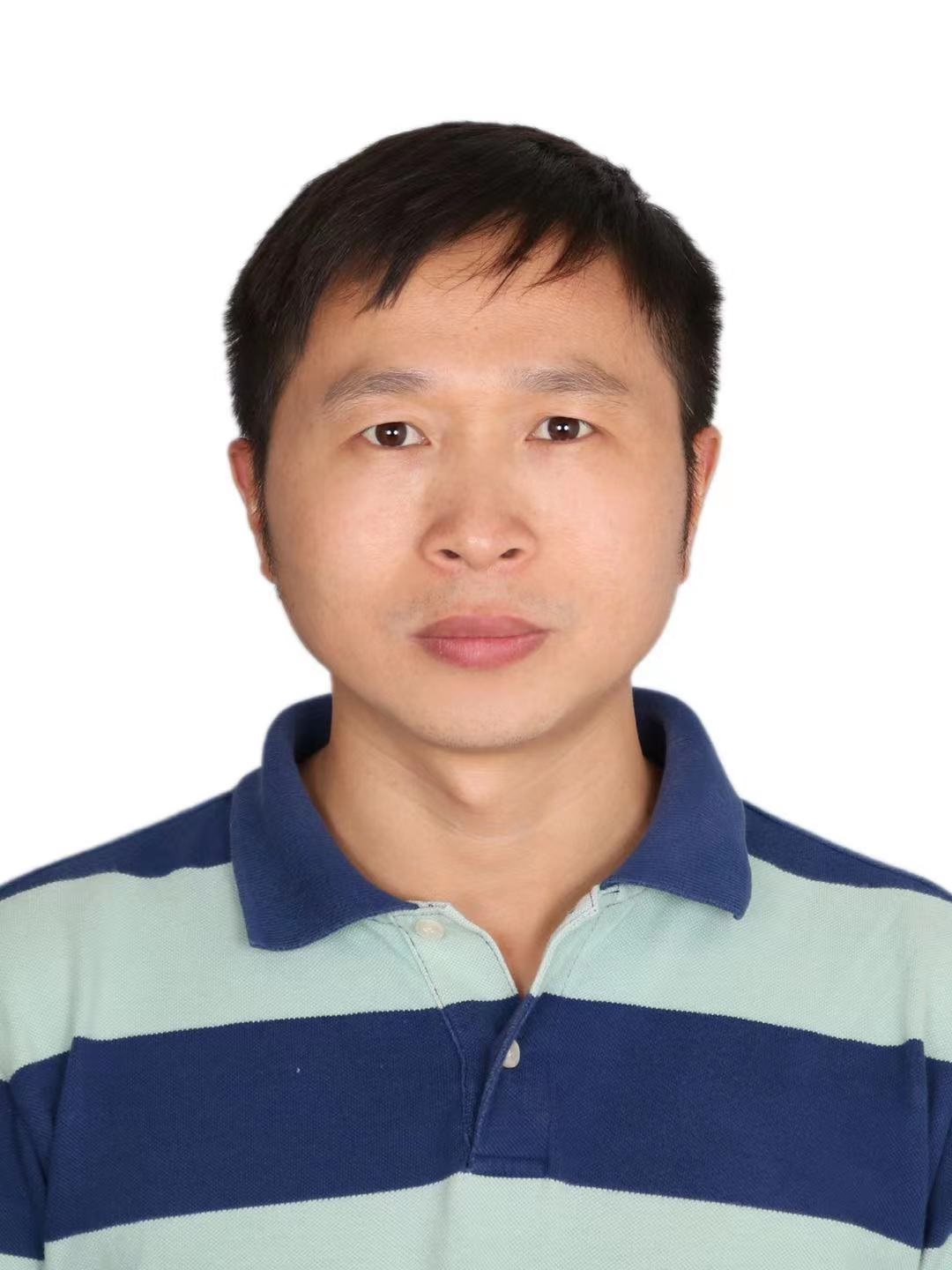}}]{Jingdong Wang} is Chief Scientist for computer vision with Baidu. Before joining Baidu, he was a Senior Principal Researcher at Microsoft Research Asia from September 2007 to August 2021. His areas of interest include vision foundation models, self-supervised pretraining, OCR, human pose estimation, semantic segmentation, image classification, object detection, and large-scale indexing. His representative works include high-resolution network (HRNet) for generic visual recognition, object-contextual representations (OCRNet) for semantic segmentation discriminative regional feature integration (DRFI) for saliency detection, neighborhood graph search (NGS, SPTAG) for vector search. He has been serving/served as an Associate Editor of IEEE TPAMI, IJCV, IEEE TMM, and IEEE TCSVT, and an (senior) area chair of leading conferences in vision, multimedia, and AI, such as CVPR, ICCV, ECCV, ACM MM, IJCAI, and AAAI. He was elected as an ACM Distinguished Member, a Fellow of IAPR, and a Fellow of IEEE, for his contributions to visual content understanding and retrieval.
\end{IEEEbiography}

\end{document}